\documentclass[10pt]{article} 
\usepackage[accepted]{tmlr}

\usepackage{hyperref}
\usepackage{url}
\usepackage{xspace}
\usepackage[pdftex]{graphicx}
\usepackage{amsmath}
\usepackage{amssymb}
\usepackage{booktabs}
\makeatletter
\DeclareRobustCommand\onedot{\futurelet\@let@token\@onedot}
\def\@onedot{\ifx\@let@token.\else.\null\fi\xspace}
\def\eg{\emph{e.g}\onedot} 
\def\ie{\emph{i.e}\onedot}

\def\wrt{w.r.t\onedot} 
\def\iid{i.i.d\onedot} 

\makeatother
\usepackage[capitalize]{cleveref}
\crefname{section}{Sec.}{Secs.}
\Crefname{section}{Section}{Sections}
\Crefname{table}{Table}{Tables}
\crefname{table}{Tab.}{Tabs.}
\usepackage{multirow}
\usepackage{algorithm}
\usepackage{algcompatible}
\newcommand{\OurModel}{DiffuseSG\xspace}
\newcommand{\OurModelnoIOU}{DiffuseSG$^{-}$\xspace}
\newcommand{\OurModelRandomNode}{DiffuseSG$^{*}$\xspace}
\newcommand{\GroundedSceneGraphs}{Grounded Scene Graphs}
\newcommand{\GroundedSceneGraph}{Grounded Scene Graph}
\newcommand{\Groundedscenegraphs}{Grounded scene graphs}
\newcommand{\Groundedscenegraph}{Grounded scene graph}
\newcommand{\groundedscenegraphs}{grounded scene graphs}
\newcommand{\groundedscenegraph}{grounded scene graph}

\newcommand{\psigma}{p_{\sigma}}
\newcommand{\R}{\mathbb{R}}

\newcommand{\btx}{\boldsymbol{\tilde{x}}}

\newcommand{\bhB}{\boldsymbol{\hat{B}}}
\newcommand{\btS}{\boldsymbol{\tilde{S}}}
\newcommand{\btQ}{\boldsymbol{\tilde{Q}}}

\newcommand{\btE}{\boldsymbol{\tilde{E}}}
\newcommand{\bte}{\boldsymbol{\tilde{e}}}
\newcommand{\btV}{\boldsymbol{\tilde{V}}}
\newcommand{\btv}{\boldsymbol{\tilde{v}}}
\newcommand{\beps}{\boldsymbol{\epsilon}}
\newcommand{\bb}{\boldsymbol{b}}

\newcommand{\bv}{\boldsymbol{v}}
\newcommand{\bx}{\boldsymbol{x}}
\newcommand{\bw}{\boldsymbol{w}}
\newcommand{\bB}{\boldsymbol{B}}
\newcommand{\bS}{\boldsymbol{S}}
\newcommand{\bE}{\boldsymbol{E}}

\newcommand{\bV}{\boldsymbol{V}}

\newcommand{\bI}{\boldsymbol{I}}
\newcommand{\bId}{\boldsymbol{I_d}}

\newcommand{\cE}{\mathcal{E}}

\newcommand{\cN}{\mathcal{N}}

\newcommand{\cS}{\mathcal{S}}
\newcommand{\cV}{\mathcal{V}}
\newcommand{\cX}{\mathcal{X}}
\newcommand{\nbyn}{n\times n}

\newcommand{\bwhS}{\boldsymbol{\widehat{S}}}
\newcommand{\bhV}{\boldsymbol{\hat{V}}}

\newcommand{\bhE}{\boldsymbol{\hat{E}}}

\newcommand{\bwtS}{\boldsymbol{\widetilde{S}}}

\title{Joint Generative Modeling of \GroundedSceneGraphs{} and Images via Diffusion Models}

\author{\name Bicheng Xu$^*$ \email bichengx@cs.ubc.ca \\
      \addr University of British Columbia \\ Vector Institute for AI
      \AND
      \name Qi Yan$^*$ \email qi.yan@ece.ubc.ca \\
      \addr University of British Columbia \\ Vector Institute for AI
      \AND
      \name Renjie Liao \email rjliao@ece.ubc.ca \\
      \addr University of British Columbia \\ Vector Institute for AI \\ Canada CIFAR AI Chair
      \AND
      \name Lele Wang \email lelewang@ece.ubc.ca \\
      \addr University of British Columbia
      \AND
      \name Leonid Sigal \email lsigal@cs.ubc.ca \\
      \addr University of British Columbia \\ Vector Institute for AI \\ Canada CIFAR AI Chair}

\def\month{08}  
\def\year{2025} 
\def\openreview{\url{https://openreview.net/forum?id=2cxxZI2LOL}} 

\begin{document}

\maketitle
\def\thefootnote{*}\footnotetext{Equal contribution}
\def\thefootnote{\arabic{footnote}}

\begin{abstract}
A \groundedscenegraph{} represents a visual scene as a graph, where nodes denote objects (including labels and spatial locations) and  directed edges encode relations among them. In this paper, we introduce a novel framework for joint \groundedscenegraph{} - image generation, a challenging task involving high-dimensional, multi-modal structured data. To effectively model this complex joint distribution, we adopt a factorized approach: first generating a \groundedscenegraph{}, followed by image generation conditioned on the generated \groundedscenegraph{}. While conditional image generation has been widely explored in the literature, our primary focus is on the generation of \groundedscenegraphs{} from noise, which provides efficient and interpretable control over the image generation process. This task requires generating plausible \groundedscenegraphs{} with heterogeneous attributes for both nodes (objects) and edges (relations among objects), encompassing continuous attributes (\eg, object bounding boxes) and discrete attributes (\eg, object and relation categories). To address this challenge, we introduce \OurModel, a novel diffusion model that jointly models the heterogeneous node and edge attributes. We explore different encoding strategies to effectively handle the categorical data. Leveraging a graph transformer as the denoiser, \OurModel progressively refines \groundedscenegraph{} representations in a continuous space before discretizing them to generate structured outputs. Additionally, we introduce an IoU-based regularization term to enhance empirical performance. Our model outperforms existing methods in \groundedscenegraph{} generation on the Visual Genome and COCO-Stuff datasets, excelling in both standard and newly introduced metrics that more accurately capture the task's complexity. Furthermore, we demonstrate the broader applicability of \OurModel in two important downstream tasks: (1) achieving superior results in a range of \groundedscenegraph{} completion tasks, and (2) enhancing \groundedscenegraph{} detection models by leveraging additional training samples generated by \OurModel. Code is available at \url{https://github.com/ubc-vision/DiffuseSG}.
\end{abstract}

\section{Introduction} \label{sec:intro}
A \groundedscenegraph{} is a graph-based representation that captures semantics of the visual scene, where nodes correspond to the objects (including their identity/labels and spatial locations) and directed edges correspond to the spatial and functional relations between pairs of objects. 
\Groundedscenegraphs{} have been widely adopted in a variety of high-level tasks, including image captioning~\citep{yang2019auto,zhong2020comprehensive} and visual question answering~\citep{damodaran2021understanding,qian2022scene}. 
Various models \citep{kundu2023ggt,jung2023devil,zheng2023prototype,jin2023fast,biswas2023probabilistic,li2024leveraging,hayder2024dsgg} have been proposed to detect 
\groundedscenegraphs{} from images. 
Such models require supervised training with image - \groundedscenegraph{} pairs, which is costly to annotate. 

\begin{figure}[t]
  \centering
   \includegraphics[width=\linewidth]{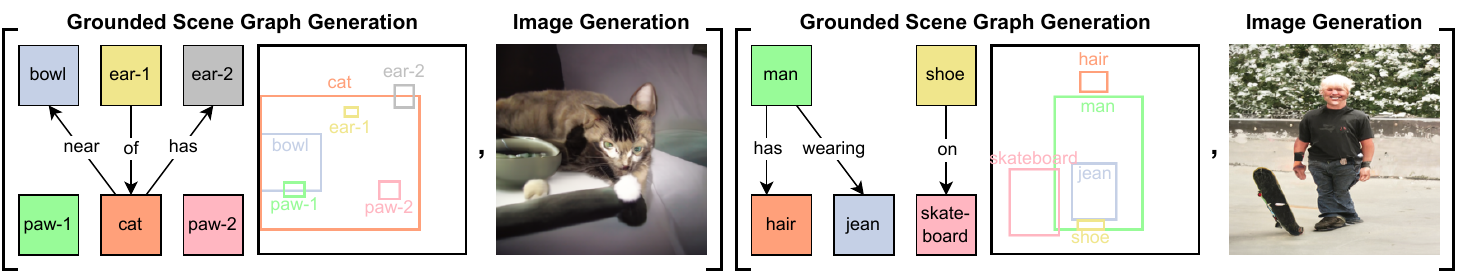}
   \caption{{\bf Joint \groundedscenegraph{} and image modeling.} We model the joint distribution of \groundedscenegraph{} - image pairs via two steps: first training our proposed \OurModel model to produce \groundedscenegraphs{} and then utilizing a  conditional image generation model to generate images. LayoutDiffusion~\citep{zheng2023layoutdiffusion} is used to generate images in these examples. Results shown are sampled from models trained on the Visual Genome dataset~\citep{krishna2017visual}.}
   \label{fig:teaser}
\end{figure}

Motivated by this and the recent successes of diffusion models, 
in this paper, we tackle the problem of joint generative modeling of \groundedscenegraphs{} and corresponding images via diffusion models. The benefits of such generative modeling would be multifaceted. First, it can be used to generate synthetic training data to augment training of discriminative \groundedscenegraph{} detection approaches discussed above. Second, it can serve as a generative scene prior which can be tasked with visualizing likely configurations of objects in the scene conditioned on partial observations via diffusion guidance. For example, where is the likely position of the chair given placement of the table and sofa. 
Third, it can be used for controlled image generation, by allowing users to sample and edit \groundedscenegraphs{}, and then conditioned on them, generate corresponding images.

Diffusion models~\citep{sohl2015deep,ho2020denoising,song2021scorebased} 
have been shown to excel at modeling complex distributions, generating realistic high-resolution images \citep{saharia2022photorealistic,rombach2022high,nair2023steered,ruiz2023dreambooth,epstein2023diffusion} and graphs \citep{jo2022score,vignac2022digress,yan2023swingnn}. 
However, joint generation of an image and the corresponding graph representation is challenging. 
To simplify the task, we factorize the joint distribution as a product of a {\groundedscenegraph{} prior} and a conditional distribution of images given a \groundedscenegraph. The conditional distribution has been widely studied in the form of {layout-to-image generation}~\citep{zhao2019image,sun2019image,sun2021learning,zheng2023layoutdiffusion}, a task of generating images based on spatial layouts that can be constructed from \groundedscenegraphs\footnote{{Though, notably, in the process, edge information is often not utilized.}}. Therefore, in this paper we mainly focus on modeling the first term, \ie, building a generative \groundedscenegraph{} model. This task in itself is challenging as it requires generation of graphs with heterogeneous attributes, \eg, real-valued node attributes like object bounding boxes and categorical edge attributes like relation types.

To generate \groundedscenegraphs, 
we propose \OurModel which is a diffusion-based generative model capable of generating plausible \groundedscenegraphs. 
To deal with heterogeneous attributes, we explore various encodings of categorical node/edge representations. 
We also design a graph transformer architecture that successively denoises the continuous graph representation which in the end produces clean \groundedscenegraph{} samples via simple discretization. 
Moreover, we introduce an intersection-over-union (IoU)-based training loss to better capture the distribution of bounding box locations and sizes. 
To generate images conditioned on \groundedscenegraphs, besides using one existing layout-to-image generation model, we also experiment with a relation-aware image generator built upon ControlNet~\citep{zhang2023adding}. 

In summary, our contributions are as follows. 
(1) We propose a novel joint \groundedscenegraph{} - image generation framework, by factorizing the joint distribution into a \groundedscenegraph{} prior and a conditional distribution
of images given the \groundedscenegraph. In this context, 
we propose a diffusion-based model, named \OurModel, for \groundedscenegraph{} generation, which jointly models 
node attributes like object classes and bounding boxes, and edge attributes like object relations.
Particularly, the technical novalties lie in the tensor representation of the \groundedscenegraph{}, separate prediction heads for node and edge attributes, an IoU-based training loss, and exploration among different encoding mechanisms for the categorical data.
(2) We show that our model significantly outperforms existing unconditional scene graph generation models, layout generation models, and general-purpose graph generative models on both standard and newly introduced metrics that better measure the similarity between observed and generated \groundedscenegraphs. 
(3) We show that our model performs well on various \groundedscenegraph{} completion tasks using diffusion guidance.
Moreover, paired with a conditional image generation model, our model generates \groundedscenegraph{} - image pairs which serve as extra training data for the downstream \groundedscenegraph{} detection task. 
The observed performance improvement highlights the practical significance of our  
joint modeling framework in real-world applications.

\section{Representation Taxonomy}
There are three similar but distinct visual scene representations that are often referenced in the literature. We formally define them here and in \cref{tab:taxonomy} to avoid conflating them. We strictly follow these definitions for the remainder of the paper.

{\bf Layout.} 
Layouts comprise objects and their corresponding bounding boxes. There is no encoding of semantic relations between these objects beyond those implied by their placement. 
This representation is widely explored in the {\em layout generation} task, and utilized in the task of {\em image generation from layouts}.

{\bf Scene Graph.} Scene graphs encode object labels and their corresponding relations in the form of directed graphs. As such, a scene graph can encode that a certain object ({\em e.g.}, a car) is {\em next to} another object, but it can not precisely localize those objects in a scene. 
This representation is mainly used in the {\em scene graph-to-image generation} task. 
More closely related to our task, some works generate scene graphs unconditionally.

{\bf \GroundedSceneGraph.} \Groundedscenegraphs{} can be viewed as a combination of 
layouts and scene graphs as defined above. A \groundedscenegraph{} is a directed graph 
where nodes correspond to objects encoded by labels \underline{and} bounding boxes, and edges correspond to relations. 
This representation is extensively studied in the literature on {\em \groundedscenegraph{} detection from images}. 
In this paper, we propose \OurModel for the task of unconditional \groundedscenegraph{} generation.

\begin{table}[t]
    \caption{\textbf{Characterization of related, but often conflated, concepts encountered in the literature.} 
    The representations used to describe the visual scene are: layout, scene graph, and \groundedscenegraph.}
    \centering
    \begin{tabular}{l|ccc}
    \toprule
    & \multicolumn{3}{c}{Attributes Contained}\\
    Concept & Object Label & Object Location & Relation Label \\
    \midrule
    Layout & \checkmark & \checkmark & \\
    Scene Graph & \checkmark & & \checkmark \\
    Grounded Scene Graph & \checkmark & \checkmark & \checkmark \\
    \bottomrule
    \end{tabular}
    \label{tab:taxonomy}
\end{table}

\section{Related Works} \label{sec:related}
{\bf Diffusion Models.} 
Diffusion models achieve great success in a variety of generation tasks nowadays, ranging from image generation~\citep{kumari2023multi,kim2024arbitrary,miao2024training,zeng2024jedi,qu2024discriminative,lin2025accdiffusion,wei2025powerful}, video generation~\citep{sun2024glober,zeng2024make,qing2024hierarchical,wang2024microcinema,skorokhodov2024hierarchical,liang2025movideo,gupta2025photorealistic,melnik2024video, liu2024ntire}, text generation~\citep{li2022diffusion,gong2022diffuseq,yuan2022seqdiffuseq,dieleman2022continuous,ye2023dinoiser,lin2023text,wu2023ar}, to simple graph generation~\citep{jo2022score,vignac2022digress,jo2023graph,yan2023swingnn,cho2024multi,xu2024discrete}, \eg, on molecule datasets~\citep{irwin2012zinc,ramakrishnan2014quantum}, which usually have less than $10$ node types and less than $5$ edge types. 
These models contain two key processes: a forward process, which typically involves adding Gaussian noise to clean data, and a reverse denoising process that is often implemented using architectures such as U-Net~\citep{ronneberger2015u} or transformer~\citep{vaswani2017attention}. 
Since the \groundedscenegraph{} is fundamentally a graph structure, our proposed \groundedscenegraph{} generation model is conceptually similar to the ones for simple graph generation. However, our model goes beyond merely generating the graph structure of node (object) and edge (relation) labels. It also generates the object locations, in the form of bounding boxes. 
Also, our \groundedscenegraph{} data is more complex than the molecule data in terms of the numbers of node and edge types, \eg, the Visual Genome dataset~\citep{krishna2017visual} contains $150$ node and $50$ edge types.
This diversity necessitates a re-evaluation of many design choices traditionally made in graph generation models.

{\bf Layout Generation.} 
Layout generation focuses on creating image layouts, which comprise object labels and their corresponding bounding box locations.
In contrast, our proposed \groundedscenegraph{} generation goes a step further by also generating the relations among these objects.
Existing layout generation models typically take the form of VAE~\citep{jyothi2019layoutvae,lee2020neural,arroyo2021variational}, GAN~\citep{li2020layoutgan,li2020attribute}, transformer or BERT type language models~\citep{gupta2021layouttransformer,kikuchi2021constrained,kong2022blt,jiang2023layoutformer++}, or diffusion models~\citep{inoue2023layoutdm,chai2023layoutdm,hui2023unifying,levi2023dlt,zhang2023layoutdiffusion,shabani2024visual}. These layout generation models usually work on graphical layout generation problems, \eg, designing layouts for mobile applications~\citep{deka2017rico,liu2018learning}, documents~\citep{zhong2019publaynet}, or magazines~\citep{zheng2019content}. The object bounding boxes of these layouts are expected to be well-aligned and not overlapping with each other. Thus, the layout generation models are usually measured on the alignment and intersection area of the generated bounding boxes. However, the object bounding boxes in our \groundedscenegraphs{} are naturally not aligned and usually occlusions occur. Therefore, we replace the evaluation metrics used in the layout generation literature with 
new ones that better capture characteristics of \groundedscenegraphs.

{\bf Unconditional Scene Graph Generation.} 
\citet{garg2021unconditional} introduce the task of unconditional scene graph generation, where a scene graph is generated from noise. They propose an autoregressive model for the generation, first sampling an initial object, then generating the scene graph in a sequence of steps. Each step generates one object node, followed by a sequence of edges connecting to the existing nodes. A follow-up work~\citep{verma2022varscene} proposes a variational autoencoder for this task, where a scene graph is viewed as a collection of star graphs. During generation, it first samples the pivot graph, and keeps adding star graphs to an existing set. The scene graph in these works is defined to only have object and relation category labels; object bounding box locations are omitted. Compared with these works, we use a diffusion-based model for unconditional \groundedscenegraph{} generation, where node and edge attributes are generated at the same time. We also include the object bounding box location in our \groundedscenegraph{} representation. 

{\bf Image Generation from Layouts or Scene Graphs.}
There exist two conditional image generation tasks in the literature: layout-to-image~\citep{zhao2019image} and scene graph-to-image~\citep{johnson2018image}. The conditions in the layout-to-image task are object labels and their bounding box locations, while the conditions in the task of scene graph-to-image are the object labels and relation labels. These tasks were initially widely explored within the GAN framework~\citep{li2019pastegan,ashual2019specifying,sun2019image,dhamo2020semantic,li2021image,sun2021learning,he2021context,sylvain2021object}. With the increasing popularity of diffusion models and their excellence at modeling complex distributions, diffusion models now become the default choice to accomplish these tasks~\citep{yang2022diffusion,cheng2023layoutdiffuse,zheng2023layoutdiffusion,farshad2023scenegenie,liu2024r3cd,liu2024hico,wang2024scene}. However, the \groundedscenegraphs{} considered in our task contain object labels, object locations, and relation labels. These attributes can not be encoded altogether into any of the existing image generation models. This motivates us to build and explore diffusion-based relation-aware layout-to-image model.

\section{Joint \GroundedSceneGraph{} - Image Pair Modeling} \label{sec:task}
We propose to model the joint distribution of \groundedscenegraphs{} and their corresponding images by modeling it as a product of two distributions. 
Denote the \groundedscenegraph{} by $\bS$ and the image by $\bI$\footnote{In what follows, we use $\bId$ for identity matrix and $\bI$ for image data.}.
The joint distribution of \groundedscenegraph{} and image pairs can be factorized as 
$p(\bS,\bI) = p(\bS) p(\bI | \bS)$, from which one can easily draw samples in a two-step manner; first from the prior $p(\bS)$ and then from the conditional $p(\bI | \bS)$.
Hence, we first build a \groundedscenegraph{} generation model to learn $p(\bS)$, \ie, the underlying prior of \groundedscenegraphs. 
Second, we employ a conditional image generation model 
to capture $p(\bI | \bS)$, the conditional image distribution. 

\subsection{\GroundedSceneGraph{} Generation}
In this section, we first formally define our proposed \groundedscenegraph{} generation task. We then present our \OurModel model specifically designed for the task. \OurModel is a continuous diffusion model, whose training and sampling utilize the stochastic differential equation (SDE) formulation. We begin by providing some background on the SDE-based diffusion modeling, followed by an in-depth explanation of \OurModel.

\subsubsection{\GroundedSceneGraph{} Generation Task}
A \groundedscenegraph{} $\bS$, consisting of $n$ nodes, can be described using node and edge tensors, denoted as $(\bV, \bE)$.
We denote the space of node features by $\cV$ and the space of edge features by $\cE$, and the space of \groundedscenegraphs{} by $\cS = \cV \times \cE$.
The node tensor $\bV = [\bv_1; \bv_2; \dots; \bv_n]\in \R^{n \times d_v}$, captures the node labels and their bounding box locations, where $d_v$ represents the dimension of the node feature. Each node feature $\bv_i = [c_i, \bb_i]$ combines a discrete node label, $c_i \in \{1, 2, \dots, Z_v\}$, with a normalized bounding box position, $\bb_i \in [0, 1]^4$.
The bounding box $\bb_i$ is represented by $(\mathrm{center}_x, \mathrm{center}_y, \mathrm{width}, \mathrm{height})$ and normalized \wrt the image canvas size.
The edge tensor $\bE \in \R^{n\times n}$, details the directed edge relationships among the nodes. Each edge entry $e_{i,j}$ corresponds to a discrete relation label, $e_{i,j} \in \{0, 1, \dots, Z_e\}$, clarifying the connections between nodes. The symbols $Z_v$ and $Z_e$ represent the total numbers of semantic object categories and relation categories of interest, respectively. Notably, $e_{i,j}=0$ indicates the absence of a relation between nodes $i$ and $j$.
The task is to generate such \groundedscenegraphs{} from noise.

\subsubsection{Diffusion Model Basics} \label{sec:sde_back}
\textbf{Preliminaries.}
Diffusion models~\citep{ho2020denoising, song2021scorebased} learn a probabilistic distribution $p_\theta(\bx)$\footnote{We use symbols $\bx, \btx$ in this section to introduce preliminaries of diffusion model in general, regardless of the type of data being modeled.} through matching the score functions of the Gaussian noise perturbed data distribution $\nabla_{\bx} \log p_\sigma(\bx)$ at various noise levels $\sigma \in \{\sigma_i\}_{i=1}^T$.
Following~\citet{song2021scorebased, karras2022elucidating}, we use the SDE-based diffusion model for training and sampling, which comes with a continuous time $t \in [0, T]$ specified by the following dynamics:
\begin{align}
    &d\bx_+ = f(\bx, t)dt + g(t) d\bw,
    \label{eq:sde_forward}
    \\
    &d\bx_- = [f(\bx, t)dt - g(t)^2 \nabla_{\bx} \log p_{t}(\bx)]dt + g(t) d\bw,
    \label{eq:sde_backward}
\end{align}
where \cref{eq:sde_forward} and \cref{eq:sde_backward} denote forward and reverse SDEs, $f(\bx, t)$ and $g(t)$ are the drift and diffusion coefficients, and $\bw$ is the standard Wiener process.
The SDEs govern how the probabilistic distribution $p_t(\bx)$ evolves \wrt time $t$.
Specifically, $p_0(\bx)$ is the data distribution, from which we observe a set of \iid samples $\cX=\{\bx_i\}_{i=1}^m$.
$p_T(\bx)$ models a tractable prior distribution, \ie, Gaussian, from which we can draw samples efficiently. 
In our formulation, we choose linear noise schedule $\sigma(t) = t$, and let $f(\bx, t) = \boldsymbol{0}$ and $g(t) = \sqrt{2\dot{\sigma}(t)\sigma(t)}$.
The SDEs solution yields that $p_t$ in~\cref{eq:sde_backward} becomes $p_{\sigma}(\bx) = p_{t}(\bx) =\frac{1}{m} \sum_{i=1}^{m} \cN(\bx; \bx_i, \sigma^2\bId)$, $\bx_i \in \cX$.
Let $p_{\cX} (\bx) = \frac{1}{m} \sum_{i=1}^{m} \delta(\bx - \bx_i)$ be the Dirac delta distribution for $\cX$.
We can rewrite $p_{\sigma}$ as $\psigma(\btx) = \int p_\cX (\bx) \psigma(\btx|\bx)d\bx $ with a Gaussian perturbation kernel $\psigma(\btx|\bx) = \cN(\btx; \bx, \sigma^2\bId)$.

\textbf{Training.}
We train a neural network to learn the score function of $p_{\sigma}$ (\ie, $\nabla_{\btx} \log \psigma(\btx)$).
Following~\citet{vincent2011connection,karras2022elucidating}, we reparameterize the score function by denoising function $D(\btx, \sigma)$ which maps the noise-corrupted data $\btx$ back to the clean data $\bx$.
They are connected by Tweedie's formula~\citep{efron2011tweedie}, 
$\nabla_{\btx} \log \psigma(\btx) = (D(\btx, \sigma) - \btx) / \sigma^2$.
In practice, we train a denoiser $D_\theta(\btx, \sigma)$ to implicitly capture the score function.
Given a specified distribution over the noise level, denoted as $p(\sigma)$, which also corresponds to the distribution of forward time $t$ since $\sigma(t) = t$, the overall training objective can be formulated as follows, 
\begin{equation}    
    \mathbb{E}_{p(\sigma) p_\cX (\bx) \psigma(\btx|\bx)}
    \left[ \Vert D_\theta(\btx, \sigma) - \bx \Vert^2_2 \right].
    \label{eq:obj_denoiser}
\end{equation}

\textbf{Sampling.}
To draw samples using the learned diffusion model, we discretize the reverse-time SDE in~\cref{eq:sde_backward} and conduct numerical integration, which gradually transitions samples from prior distribution $p_T$ to data distribution $p_0$. We choose a set of discrete time steps $\{t_i\}_{i=1}^T$, at which the score function is evaluated using the trained model for numerical reverse-SDE solution. We employ a second-order solver based on Heun's method~\citep{suli2003introduction,karras2022elucidating}.

\begin{figure*}[t]
  \centering
   \includegraphics[width=\linewidth]{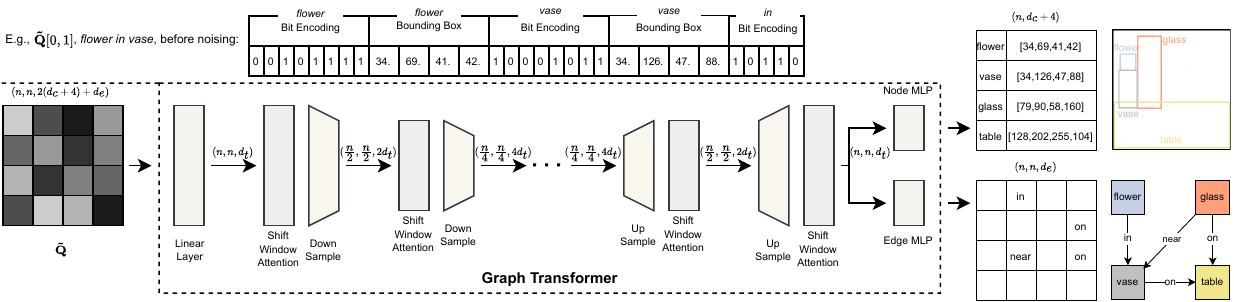}
   \caption{{\bf Illustration of the graph transformer during training.} Object and relation categories are illustrated using binary-bit encoding. Object bounding box positions are represented by $(\mathrm{center}_x, \mathrm{center}_y, \mathrm{width}, \mathrm{height})$.}
   \label{fig:diffusegsg}
\end{figure*}

\subsubsection{\OurModel for \GroundedSceneGraph{} Generation} \label{sec:model}
We 
model the \groundedscenegraph{} distribution $p(\bS)$ with a continuous state diffusion model.
Our model captures the distribution of \groundedscenegraph{} topology along with node and edge attributes simultaneously. 

\textbf{Denoising Objective.}
During training, we draw noisy \groundedscenegraph{} samples from the perturbed distribution $\psigma(\btS) = \int p_\cS (\bS) \psigma(\btS|\bS)d\bS $ and train a denoiser $D_\theta(\btS, \sigma)$ to output the associated noise-free samples $\bS$.
We relax node and edge label distributions to continuous space, enabling SDE-based diffusion modeling with smooth Gaussian noise on all attributes.
Various methods for encoding discrete labels will be introduced in Section ``Encoding Discrete Data'' below.
Specifically, $\btS = (\btV, \btE)$ denotes the noisy \groundedscenegraph{} and $p_\cS (\bS) = \frac{1}{m} \sum_{i=1}^{m} \delta(\bS - \bS_i)$ is the Dirac delta distribution based on training data $\{\bS_i\}_{i=1}^m$.
We implement the \groundedscenegraph{} Gaussian perturbation kernel  $\psigma(\btS|\bS)$ by independently injecting noise to node and edge attributes, \ie, 
$\psigma(\btS)= \int p_\cS (\bS) \psigma(\btV|\bV) \psigma(\btE|\bE) d\bS$.
The decomposed kernels are both simple Gaussians: $\psigma(\btV|\bV) = \cN(\btV; \bV, \sigma^2 \bId)$, $\psigma(\btE|\bE) = \cN(\btE; \bE, \sigma^2 \bId)$. 
Further, we design a denoising network $D_\theta$ with two prediction heads $D_\theta^V$ and $D_\theta^E$ dedicated to node and edge attributes respectively (detailed in Section ``Network Design'' below). 
Our \groundedscenegraph{} denoising loss now becomes:
\begin{align}
    \mathcal{L}_{d} = 
    \mathbb{E}_{p(\sigma) p_\cS (\bS) \psigma(\btS|\bS)}
    [ \Vert D_\theta^V(\btS, \sigma) - \bV \Vert^2_2 + \Vert D_\theta^E(\btS, \sigma) - \bE \Vert^2_2].
    \label{eq:obj_sg_denoiser}
\end{align}

\textbf{Network Design.}
To effectively capture the complex distribution of \groundedscenegraphs, we develop a  
transformer architecture, named 
{\em graph transformer}, as the denoiser $D_\theta$. 
Given a noisy \groundedscenegraph{} $\btS = (\btV, \btE)$ as input, we construct triplet representations (\ie, generalized edge representations) by concatenating the subject node, object node, and relation information $\btQ[i,j] = [\btv_i, \btv_j, \bte_{i, j}], \forall i,j \in [n]$, where $n$ is the number of nodes in the \groundedscenegraph{}. The denoising task is essentially node and edge regression in continuous space, trained with stochasticity. 
For expressive graph representation learning in this context, we consider 
message passing among all $O(n^2)$ triplets as suggested in~\citet{morris2021weisfeilera,morris2021weisfeiler}. Here, each triplet becomes a unit of message passing. 
However, a naive triplet-to-triplet message passing implementation is space-consuming ($O(n^4)$ messages). Inspired by~\citet{liu2021swin}, we employ an approximate triplet-to-triplet message passing using shifted-window attention layers with a window size $M$, which reduces the space complexity to $O(n^2M^2)$. When window-partitioning is repeated adequately, \eg, at least $O(n/M)$ times, all triplet-to-triplet interactions can be effectively approximated.
We tokenize the noisy triplets using a linear layer on each entry in $\btQ$, resulting in $\nbyn$ triplet tokens of dimension $d_t$, represented as $\btQ_d \in \R^{\nbyn \times d_t}$. Our graph transformer then employs repeated shifted-window attention and downsampling/upsampling layers to update the dense triplet-token representations, for predicting the noise-free node and edge attributes.
To generate node and edge attribute predictions of distinct shapes, we employ two MLPs as readout layers for node and edge respectively.
The node denoiser $D_\theta^V$ and the edge denoiser $D_\theta^E$ share identical intermediate feature maps and have the same parameters up to the final readout layers. 
The network architecture is further detailed in~\cref{app:diffusesg_net}.

\textbf{Encoding Discrete Data.}
To find an appropriate representation for the categorical labels, denoting the node label representation as $c'_i \in \R^{d_{c}}$ and the edge label representation as $e'_{i,j} \in \R^{d_{e}}$, we explore three distinct encoding methods.
(1) Scalar: both node and edge labels are expressed as scalar values, with $d_c = d_e = 1$.
(2) Binary-Bit Encoding: the discrete type indices of node and edge labels are converted into their binary format, represented as a sequence of $0$s and $1$s. Here, $d_c = \lceil \log_2(Z_v) \rceil$ and $d_e = \lceil \log_2(Z_e+1) \rceil$. 
(3) One-Hot Encoding: the scalar labels are transformed into their one-hot representations, resulting in $d_c = Z_v$ 
and $d_e = Z_e+1$. 
The impact of these different encoding methods is further analyzed and compared in~\cref{sec:sgg_exp}.
During training, after encoding the node and edge labels, along with the bounding box positions, we inject noise and form the noisy $\btQ$.
\cref{fig:diffusegsg} illustrates the pipeline of our graph transformer during training.
During sampling, we start with a Gaussian noised $\btQ$, where the node number can either be specified or drawn from some given distribution. After the denoising process, we discretize the continuous-valued representations of node and edge types, 
\eg, through thresholding for the binary-bit encoding. Detailed explanations of these categorical encodings and discretizations during sampling are in~\cref{app:input_code}.
Note that the bounding box positions 
$\bb_i$
are naturally continuous and there is no need for 
discretization while sampling.

\begin{algorithm}[t]
\caption{\OurModel Training Process.}
\label{alg:trainer}
\begin{algorithmic}[1]
\REQUIRE{node denoiser $D_\theta^V$, edge denoiser $D_\theta^E$, diffusion time distribution $p(\sigma)$, dataset $\cX$.}
\REPEAT
\STATE {\bfseries sample} {$ \bS \sim \cX$} \hfill $\rhd$ {Draw one grounded scene graph sample}
\STATE {\bfseries sample} {$ \sigma \sim p(\sigma)$} \hfill $\rhd$ {Draw a diffusion timestamp}
\STATE {$(\bV, \bE) \leftarrow \bS$} \hfill $\rhd$ {Obtain the node and edge tensors}
\STATE {$\bB \leftarrow \bV$} \hfill $\rhd$ {Obtain the bounding box positions}
\STATE {$\btV \leftarrow \bV + \sigma \beps_1$ with $\beps_1 \sim \cN(\boldsymbol{0}, \bI)$} \hfill $\rhd$ {Add noise to the node tensor}
\STATE {$\btE \leftarrow \bE + \sigma \beps_2$ with $\beps_2 \sim \cN(\boldsymbol{0}, \bI)$} \hfill $\rhd$ {Add noise to the edge tensor}
\STATE {$\btS \leftarrow (\btV, \btE)$}
\STATE {$\bhV = D_\theta^V(\btS, \sigma)$} \hfill $\rhd$ {Denoise the node tensor}
\STATE {$\bhE = D_\theta^E(\btS, \sigma)$} \hfill $\rhd$ {Denoise the edge tensor}
\STATE {$\mathcal{L}_{d} = 
    \Vert \bhV - \bV \Vert^2_2 + \Vert \bhE - \bE \Vert^2_2$} \hfill $\rhd$ {\cref{eq:obj_sg_denoiser}}
\STATE {$\bhB \leftarrow \bhV$} \hfill $\rhd$ {Obtain the denoised bounding box positions}
\STATE {$\mathcal{L}_{iou} = 1 - \frac{1}{n} \sum_{i = 1}^{n} \operatorname{GIoU}(\bhB_i, \bB_i)$} \hfill $\rhd$ {\cref{eq:loss_iou}}
\STATE {{\bfseries update} $\theta$ via $-\nabla_\theta (\mathcal{L}_{d} + \lambda \mathcal{L}_{iou})$} \hfill $\rhd$ {Optimization step}
\UNTIL {converged}
\end{algorithmic}
\end{algorithm} 

\textbf{Additional Bounding Box IoU Loss.}
To enhance bounding box generation quality, we integrate an intersection-over-union (IoU)-based loss, $\mathcal{L}_{iou}$, into the denoising objective. This IoU loss aims to align the denoised bounding boxes $\bhB \in \mathbb{R}^{n \times 4}$ (a partial output of $D_\theta^V$) closely with the ground-truth $\bB \in \mathbb{R}^{n \times 4}$. The IoU loss is formulated as: 
\begin{equation} \label{eq:loss_iou}
\mathcal{L}_{iou} = 1 - \frac{1}{n} \sum_{i = 1}^{n} \operatorname{GIoU}(\bhB_i, \bB_i), 
\end{equation}
where $n$ is the number of objects in the \groundedscenegraph, and $\operatorname{GIoU}$ is the generalized IoU, proposed in~\citet{rezatofighi2019generalized}, of the corresponding boxes.
The final training loss then becomes:
\begin{equation} \label{eq:total_loss}
\mathcal{L} = \mathcal{L}_{d} + \lambda \mathcal{L}_{iou}, 
\end{equation}
where $\lambda$ is a hyperparameter that adjusts the balance between these two loss components. Note, $\mathcal{L}_{d}$ is given in~\cref{eq:obj_sg_denoiser}. 
We use $\lambda=1$ in our experiments.
\cref{alg:trainer} shows the training process of \OurModel. Note, the node and edge denoisers ($D_\theta^V$ and $D_\theta^E$) share identical intermediate feature maps and have the same parameters up to the final prediction heads, and the bounding box positions are part of the node tensor. The diffusion modeling details are in~\cref{app:diffusesg_detail}.

{\bf Summary of Technical Contributions.} Though the shifted window attention is adopted from~\citet{liu2021swin}, and the U-Net architecture is similar to other diffusion models like the ones in~\citet{song2021scorebased,karras2022elucidating}. Our novelties lie in: (1) the tensor representation of the grounded scene graph for \OurModel (a continuous diffusion model), (2) the separate read out layers (MLP prediction heads) for object and relation generation, (3) an additional bounding box IoU loss for training, and (4) exploration among different encoding mechanisms for object and relation categories.

\subsection{Conditional Image Generation} \label{sec:model_images}
We use two diffusion-based conditional image generators to model the conditional image distribution $p(\bI | \bS)$ given generated \groundedscenegraphs: one existing layout-to-image generator named \textit{LayoutDiffusion}~\citep{zheng2023layoutdiffusion}, and one relation-aware image generator which is built upon ControlNet~\citep{zhang2023adding} by ourselves, termed \textit{Relation-ControlNet}.

\subsubsection{LayoutDiffusion}
LayoutDiffusion~\citep{zheng2023layoutdiffusion}, is a diffusion-based layout-to-image generation model. 
It uses a U-Net architecture for the denoising process, with the layout condition enforced on the hidden features of the U-Net. It first employs a transformer-based layout fusion module to capture the information in the given layout, and then utilizes a cross-attention mechanism to fuse the image features and the layout representations inside the denoising U-Net. The whole diffusion process is applied on the image pixel space. Following~\citet{ho2020denoising,ho2022classifier}, LayoutDiffusion utilizes a standard mean-squared error loss to train the diffusion model and the classifier-free guidance technique to support the layout condition. 

LayoutDiffusion is trained and evaluated on the Visual Genome~\citep{krishna2017visual} and COCO-Stuff~\citep{caesar2018coco} datasets, which is aligned with our dataset settings as described in \cref{sec:dataset}. Given the superior performance of LayoutDiffusion on these two datasets, we decide to adopt it as one of our conditional image generator candidates. Specifically, we take the model checkpoints provided by the authors\footnote{\url{https://github.com/ZGCTroy/LayoutDiffusion}} which generate images of resolution $256 \times 256$.

\subsubsection{Relation-Aware Layout-to-Image Generation (Relation-ControlNet)}
As discussed in \cref{sec:related}, the existing conditional image generators are either conditioned only on object labels and their locations (layout-to-image models), or object labels and relation labels but no object locations (scene graph-to-image models). To fully utilize the data information generated by our \OurModel, we build a relation-aware layout-to-image generator, which generates images conditioned on object labels, locations and relation labels. We call the model, which is based on ControlNet~\citep{zhang2023adding}, Relation-ControlNet.

\begin{figure*}[t]
  \centering
   \includegraphics[width=\linewidth]{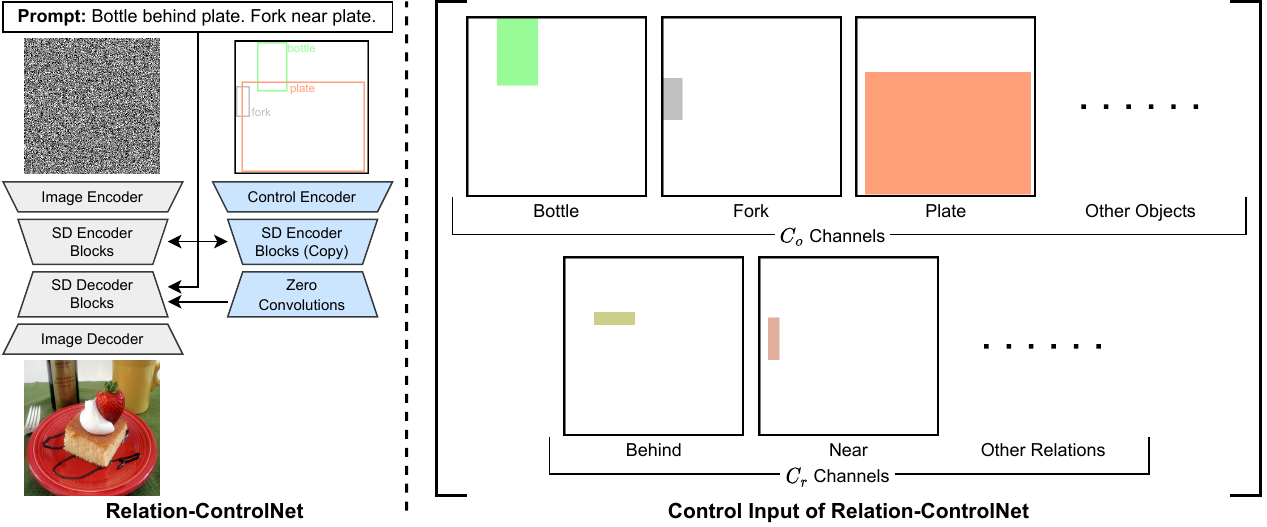}
   \caption{{\bf Relation-ControlNet Illustration.} The control input is to provide spatial information for both objects and relations. Each channel in the control input has size $(256, 256)$, where the colored areas having values of $1$ and the white areas are filled with values of $0$. In the figure, SD stands for Stable Diffusion.}
   \label{fig:relationcontrol}
\end{figure*}

\textbf{ControlNet.}
ControlNet~\citep{zhang2023adding} is a neural network architecture which can add spatial conditioning controls to large pre-trained text-to-image diffusion models. When it is instantiated on the pre-trained Stable Diffusion~\citep{rombach2022high} model, the encoder part of the U-Net is first cloned. The image feature is inputted to both the original U-Net encoder and the cloned counterpart, while the cloned network blocks receive an additional control input. The output of the cloned U-Net encoder is then injected into the original U-Net decoder via \textit{zero convolution layers}. The control input has the same spatial dimension as the input/output image. Since the U-Net in Stable Diffusion works on the latent image features, the control input is also transformed by convolution layers to match the input dimension of the diffusion U-Net. The original training loss in Stable Diffusion is used. During training, only the weights of the cloned U-Net encoder and the newly introduced convolution layers get updated. 
We utilize ControlNet with Stable Diffusion V1.5 to build our relation-aware layout-to-image generation model. Specifically, we tailor the control input and the text prompt of ControlNet to encode the object label, object location, relation label, and relation location information. We train the ControlNet to produce images of resolution $256 \times 256$.

\textbf{Control Input.} The control input to the ControlNet is to specify the spacial information. We utilize this to provide the spacial information of both objects and relations given a \groundedscenegraph. Specifically, assuming there are $C_o$ object categories and $C_r$ relation classes given a dataset, we construct a control input of size $(256, 256, C_o + C_r)$ to encode the object and relation locations, where we use one channel to represent one specific object/relation category. That is, the channel $C_i$ is to provide the location information of object/relation class $C_i$, and all objects or relations having the same class are encoded in the same channel. The first $C_o$ channels are for object classes while the last $C_r$ channels are for relation categories. The relation location is determined via the ``between'' operation given the subject and object bounding boxes as defined in~\citet{hoe2024interactdiffusion}. The bounding box positions are mapped from $[0, 1]$ in continuous space to $[0, 1, ..., 255]$ discrete grids to form the control input.

\textbf{Text Prompt.} Given the specific version of Stable Diffusion (V1.5) used, the text prompt encoder can only encode $77$ tokens. Therefore, we let the text prompt only include the relation information. We define a relation sentence as a concatenation of the subject label, the relation label, and the object label for a relation triplet, ending with a period, \eg, ``cat under chair.''. Given the limited capability of the text encoder, we decide to only encode the unique relation sentences given a \groundedscenegraph. Same as in~\citet{zhang2023adding}, during training, the text prompt is replaced with an empty string at a probability of $50\%$.

The Relation-ControlNet is illustrated in~\cref{fig:relationcontrol}, with emphasis on the control input. The training details can be found in~\cref{app:controlnet}.

\section{Experiments}
We conduct all experiments on the Visual Genome~\citep{krishna2017visual} and COCO-Stuff~\citep{caesar2018coco} datasets. 
Our experiments mainly focus on the \groundedscenegraph{} generation part since this is the main contribution of this work.

\subsection{Datasets} \label{sec:dataset}
{\bf Visual Genome (VG).} 
We use the same pre-processing procedure and train/val splits as previous \groundedscenegraph{} detection works~\citep{xu2017scene,zellers2018neural,tang2020unbiased,li2022sgtr}, but only the \groundedscenegraph{} annotations are used. This pre-processed dataset contains $57,723$ training and $5,000$ validation \groundedscenegraphs{} with $150$ object and $50$ relation categories. For each \groundedscenegraph, 
we ensure that there will be only one edge label if there exists a directed edge (relation) between two nodes (objects). 
Each \groundedscenegraph{} has node numbers between $2$ and $62$, with an average of $5.15$ relations per instance.

{\bf COCO-Stuff.} 
COCO-Stuff contains 171 object types (including 80 thing and 91 stuff categories). 
It comes with object label and bounding box annotations, but no relation labels. Following~\citet{johnson2018image}, we manually assign a relation label between two bounding boxes based on their relative positions, with a relation set of $6$ labels: {\tt left of}, {\tt right of}, {\tt above}, {\tt below}, {\tt inside}, and {\tt surrounding}. Same as in~\citet{kong2022blt}, we remove small bounding boxes ($\leq 2\%$ image area) and instances tagged as ``iscrowd'', resulting in $118,262$ training and $4,999$ validation \groundedscenegraphs. Each \groundedscenegraph{} is a fully-connected graph without self-loop, with node numbers between $1$ and $33$. 

\subsection{\GroundedSceneGraph{} Generation Experiments}
\subsubsection{Evaluation Metrics}
We use maximum mean discrepancy (MMD), triplet label total variation difference (Triplet TV), and our proposed novel object detection-based F1 scores to measure the model performance on the \groundedscenegraph{} generation task.

{\bf MMD.} 
Inspired by the relevant graph generation literature~\citep{you2018graphrnn,liao2019efficient}, we use MMDs 
to measure the similarities between the generated \groundedscenegraphs{} and the ground-truth ones on the node degree, node label, and edge label distributions respectively. Empirically, let $\{x_i\}_{i=1}^n$ be the generated samples and $\{y_j\}_{j=1}^m$ be the ground-truth ones. The MMD value is calculated as $\frac{1}{n^2} \sum_{i=1}^n \sum_{j=1}^n k(x_i, x_j) + \frac{1}{m^2} \sum_{i=1}^m \sum_{j=1}^m k(y_i, y_j) - \frac{2}{nm} \sum_{i=1}^n \sum_{j=1}^m k(x_i, y_j)$, where $n$ and $m$ are the numbers of generated samples and ground-truth ones respectively, and we use the Gaussian kernel as the kernel function $k(\cdot , \cdot)$. The lower MMD value means the closer to the ground-truth distribution.

{\bf Triplet TV.} 
As the labels of the <{\tt subject}, {\tt relation}, {\tt object}> 
triplets lie in a very high dimension 
(the cross product of potential subject, relation, and object types), 
it is computationally infeasible to calculate the triplet label MMD. As a compromise, we use the total variation difference (TV) to measure the marginal distribution difference between the generated triplet labels and the ground-truth ones. 
Specifically, assuming that the generated empirical distribution is $\hat{p}$ with the ground-truth being $\hat{q}$, where $\hat{p}$ and $\hat{q}$ are vectors sized as the number of unique triplets combining all generated and ground-truth triplets. The TV is calculated as $\frac{1}{2} |\hat{p} - \hat{q}$|. Note, if a triplet does not exist in either $\hat{p}$ or $\hat{q}$, then its relevant entry in $\hat{p}$ or $\hat{q}$ is $0$.

{\bf Detection-based F1 Scores.} We propose a set of novel object detection-based F1 scores to evaluate the generated bounding box layout quality (including both the location and node label). 
Specifically, for a generated layout, 
we calculate a F1 score between this generated layout and every ground-truth layout, and take the maximum one as the final score. 
Assume there are $N$ node categories.
Given a pair of generated and ground-truth layouts, the F1 score is calculated as
$\mathrm{F1} = \sum_{c \in N} (w_c \cdot \mathrm{F1}_{c})$,
where $\text{F1}_{c}$ is the F1 score for a node category $c$ and $w_c$ is its weighting coefficient. 
Calculating $\mathrm{F1}_{c}$ needs to decide whether two bounding boxes match or not. We use $10$ different bounding box IoU thresholds ranging from $0.05$ to $0.5$ with a step size of $0.05$ to decide the bounding box match.
That is, 
$\mathrm{F1}_{c} = \frac{1}{10} \sum_{\mathrm{iou} \in [0.05:0.05:0.5]} \mathrm{F1} (\mathrm{iou} | c)$, 
where $\mathrm{F1} (\mathrm{iou} | c)$ means a F1 score between two layouts given a specific node category $c$ and a IoU threshold $\mathrm{iou}$. 
We calculate $4$ different types of F1 scores: 
(1) F1-Vanilla (F1-V), where $w_c$ is set to $\frac{1}{|N|}$ for every node category;
(2) F1-Area (F1-A), where $w_c$ is set to $\frac{\operatorname{Area}(c)}{\sum_{c \in N}\operatorname{Area}(c)}$ and $\operatorname{Area}(c)$ is the average bounding box area in the validation set for the node category $c$;
(3) F1-Frequency (F1-F), where $w_c$ is set to $\frac{\operatorname{Freq}(c)}{\sum_{c \in N}\operatorname{Freq}(c)}$ and $\operatorname{Freq}(c)$ is the frequency of the node category $c$ in the validation set; 
(4) F1-BBox Only (F1-BO), where the F1 calculation is purely based on the bounding box locations, 
that is, we treat all bounding boxes as having a single node category ($|N| = 1$ and $w_c = 1$). 
The motivation of having F1-Area and F1-Frequency is that we want some metrics to be slightly biased to those salient objects (appearing either in a large size in general or more frequently). 
The higher the F1 scores, the better.

\subsubsection{Baselines}
We consider the following six baselines to compare to our \OurModel model.

\begin{table}[t]
    \caption{\textbf{Different attributes} that can be generated from different methods.}
    \centering
    \begin{tabular}{l|c|ccc}
    \toprule
    & Generation & \multicolumn{3}{c}{Attributes Generated}\\
    Method & Type & Object Label & Object Location & Relation Label \\
    \midrule
    SceneGraphGen & Scene Graph & \checkmark & & \checkmark \\
    VarScene & Scene Graph & \checkmark & & \checkmark \\
    D3PM & Scene Graph & \checkmark & & \checkmark \\
    BLT & Layout & \checkmark & \checkmark & \\
    LayoutDM & Layout & \checkmark & \checkmark & \\
    DiGress & Grounded Scene Graph & \checkmark & \checkmark & \checkmark \\
    \OurModel & Grounded Scene Graph & \checkmark & \checkmark & \checkmark \\
    \bottomrule
    \end{tabular}
    \label{tab:baselines}
\end{table}

(1) \textbf{SceneGraphGen}~\citep{garg2021unconditional} is a scene graph generative model based on recurrent networks. It is an autoregressive model where one object label or one relation label is generated at a time. This model is not capable of producing object bounding boxes, and we can not specify the number of objects in a scene graph to be generated during inference. 

(2) \textbf{VarScene}~\citep{verma2022varscene} is a variational autoencoder-based generative model for scene graphs. 
Similar to SceneGraphGen, it generates scene graphs without bounding boxes and can not accept number of objects as parameter during  inference. Using the released code\footnote{\url{https://github.com/structlearning/varscene/tree/main}}, we successfully replicate the results on the VG dataset but encounter issues with COCO-Stuff, as its symmetric modeling of edges between nodes prevents VarScene from generating directed graphs. Consequently, we report directed graph results on VG but undirected graph results on COCO-Stuff. 

(3) \textbf{D3PM}~\citep{austin2021structured} is a discrete denoising diffusion probabilistic framework designed for discrete data generation. We adopt its image generation model for scene graph generation; details are in~\cref{app:d3pm}. This model only generates node and edge labels. 

(4) \textbf{BLT}~\citep{kong2022blt} is a transformer-based layout generation model where only object labels and bounding boxes are generated. The transformer is non-autoregressive, where all the attributes of the layout are generated at the same time as discrete tokens. Bounding box locations are quantized into integers. 

(5) \textbf{LayoutDM}~\citep{inoue2023layoutdm} is a discrete state-space diffusion model for layout generation. Similar to BLT, this model is also not able to model relation labels and the bounding box locations are discretized.

(6) \textbf{DiGress}~\citep{vignac2022digress} is a discrete denoising diffusion model for generating graphs with categorical node and edge labels. We add an additional input of discrete bounding box representation, same as the one used in LayoutDM, to incorporate bounding box generation. 

\cref{tab:baselines} shows the different attributes that can be generated among all the baselines and \OurModel.
Among the six baselines, SceneGraphGen, VarScene, and LayoutDM can not deal with specification on the number of objects for the scene graphs or layouts to be generated during sampling, while the other three can. Comparing to those diffusion-based baselines (D3PM, LayoutDM, and DiGress), \OurModel performs the diffusion process in the continuous space. For all the baselines, we train them from scratch using the authors' released code, with slight adaptation to the datasets.

\begin{table*}[t]
    \caption{\textbf{\Groundedscenegraph{} generation results} on the Visual Genome and COCO-Stuff validation sets. In each column, the best value is \textbf{bolded}. 
    N-MMD, D-MMD, and E-MMD are the MMD values calculated based on node label distribution, node degree distribution, and edge label distribution respectively. T-TV (val) / (train) is the Triplet TV calculated against validation / training triplet statistics. The training set has a larger set of triplets than the validation, giving a more comprehensive evaluation.}
    \centering
    \resizebox{\linewidth}{!}{
    \begin{tabular}{l|ccccc|cccc}
    \toprule 
    \multicolumn{10}{c}{Visual Genome (VG)} \\
    \midrule
    Method & N-MMD$\downarrow$ & F1-V$\uparrow$ & F1-A$\uparrow$ & F1-F$\uparrow$ & F1-BO$\uparrow$ & D-MMD$\downarrow$ & E-MMD$\downarrow$ & T-TV (val)$\downarrow$ & T-TV (train)$\downarrow$ \\ 
    \midrule
    LayoutDM & 9.44e-3 & 0.161 & 0.291 & 0.368 & \textbf{0.766} & - & - & - & - \\
    SceneGraphGen & \textbf{8.77e-3} & - & - & - & - & 3.79e-2 & \textbf{2.29e-2} & 0.987 & 0.979 \\
    VarScene & 2.58e-2 & - & - & - & - & 1.04e-2 & 3.91e-2 & 0.988 & 0.981 \\
    \OurModelRandomNode & 9.52e-3 & {\bf 0.188} & {\bf 0.331} & {\bf 0.369} & {0.749} & {\bf 6.35e-3} & {3.25e-2} & \textbf{0.735} & \textbf{0.566} \\
    \midrule
    BLT & 2.70e-2 & {0.181} & {0.300} & \textbf{0.376} & 0.708  & - & - & - & - \\
    D3PM & 7.69e-3 & - & - & - & - & 3.07e-2 & {2.00e-2} & 0.816 & 0.772 \\
    DiGress & 7.94e-3 & 0.157 & 0.263 & 0.282 & 0.732 & 8.89e-3 & \textbf{8.02e-3} & 0.718 & 0.706 \\
    \OurModel & \textbf{6.64e-3} & \textbf{0.184} & \textbf{0.308} & {0.292} & \textbf{0.747} & \textbf{5.26e-3} & 3.46e-2 & \textbf{0.702} & \textbf{0.685} \\
    \bottomrule
    \toprule
    \multicolumn{10}{c}{COCO-Stuff} \\
    \midrule
    Method & N-MMD$\downarrow$ & F1-V$\uparrow$ & F1-A$\uparrow$ & F1-F$\uparrow$ & F1-BO$\uparrow$ & D-MMD$\downarrow$ & E-MMD$\downarrow$ & T-TV (val)$\downarrow$ & T-TV (train)$\downarrow$  \\ 
    \midrule
    LayoutDM & \textbf{3.40e-4} & 0.274 & 0.330 & 0.508 & {\bf 0.824} & - & - & - & - \\
    SceneGraphGen & 3.79e-4 & - & - & - & - & 2.59e-3 & 7.24e-4 & 0.904 & 0.895 \\
    VarScene & 2.60e-2 & - & - & - & - & 3.07e-1 & 9.32e-2 & 0.949 & 0.949 \\
   \OurModelRandomNode & 1.22e-3 & \textbf{0.439} & \textbf{0.500} & \textbf{0.639} & {0.822} & \textbf{1.44e-4} & \textbf{1.59e-4} & \textbf{0.229} & \textbf{0.287} \\
   \midrule
    BLT & 1.09e-1 & 0.322 & 0.389 & 0.526 & 0.807  & - & - & - & - \\
    D3PM & \textbf{4.92e-4} & - & - & - & - & 0 & 1.29e-4 & 0.341 & 0.305 \\
    DiGress & 1.06e-3 & 0.342 & 0.387 & 0.570 & 0.782 & 0 & 4.44e-3 & 0.515 & 0.398 \\
    \OurModel & {5.53e-4} & \textbf{0.421} & \textbf{0.485} & \textbf{0.637} & \textbf{0.830} & 0 & \textbf{7.25e-5} & \textbf{0.270} & \textbf{0.219} \\
    \bottomrule
    \end{tabular}
    }
    \label{tab:sgg_vg}
\end{table*}

\begin{figure*}[t]
  \centering
   \includegraphics[width=\linewidth]{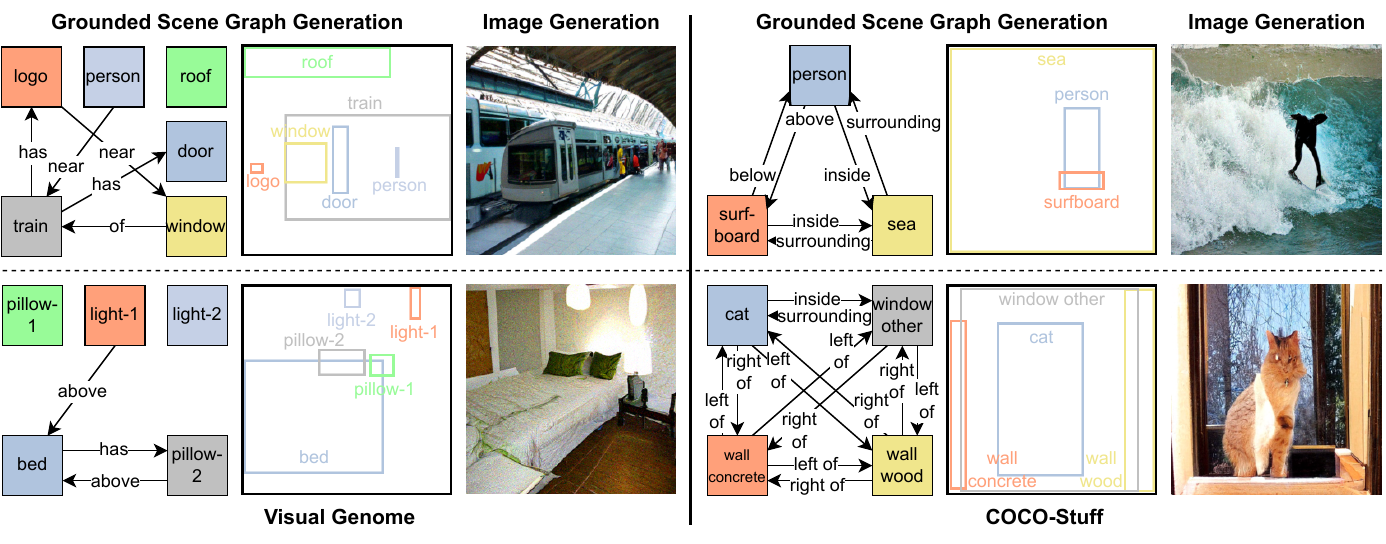}
   \caption{{\bf \Groundedscenegraph{} - image pair generation qualitative results.} \Groundedscenegraphs{} are generated by \OurModel and the corresponding $256 \times 256$ images are produced by LayoutDiffusion~\citep{zheng2023layoutdiffusion}.}
   \label{fig:sgg}
\end{figure*}

\subsubsection{\GroundedSceneGraph{} Generation Results} \label{sec:sgg_exp}
Our \OurModel is trained with~\cref{eq:total_loss} and uses the binary-bit input representation. 
To compare with SceneGraphGen, VarScene, and LayoutDM, while sampling with \OurModel, we do not fix the object numbers but draw the number of objects from its empirical distribution on the validation set. This line of results is denoted as \OurModelRandomNode. Since the other three models (D3PM, BLT, and DiGress) can specify the object numbers during inference, when comparing with these three models, the ground-truth object numbers are used for \groundedscenegraph{} generation.
For all our models and baselines, we consistently randomly sample a fixed set of $1,000$ training samples to do model selection. 

\cref{tab:sgg_vg} 
shows the quantitative results. 
For each model, we generate $3$ sets of layout, scene graph, or \groundedscenegraph{} samples, where the size of one set samples is equal to the number of instances in the respective validation set. 
Reported results are the averaged results over the $3$ sample sets. 
From the table, we can see that our \OurModelRandomNode and \OurModel achieves the best results on most evaluation metrics. 
Comparing \OurModelRandomNode with LayoutDM, SceneGraphGen, and VarScene, \OurModelRandomNode is better than these baselines on F1-V, F1-A, and F1-F scores (describing both object labels and bounding box locations), D-MMD (measuring the graph connectivity), and Triplet-TV (capturing the co-occurrence between objects and their relevant edge labels). Notably, \OurModelRandomNode is better than the second best model SceneGraphGen by $83.25\%$ (VG) and $94.44\%$ (COCO-Stuff) regarding D-MMD, and by $25.53\%$ (VG) and $74.67\%$ with respect to T-TV (val). 
\OurModelRandomNode is only worse than SceneGraphGen on N-MMD (VG) and E-MMD (VG), and LayoutDM on N-MMD (COCO-Stuff) and F1-BO (VG and COCO-Stuff); but the differences are marginal. 
Comparing \OurModel with BLT, D3PM, and DiGress, \OurModel is better than these baselines on all F1 scores (except F1-F on VG), and Triplet-TV. Remarkably, on COCO-Stuff, \OurModel is $23.10\%$, $25.32\%$, and $11.75\%$ better than DiGress with regard to F1-V, F1-A, and F1-F respectively. 
\OurModel is only worse than D3PM on N-MMD (COCO-Stuff), BLT on F1-F (VG), and DiGress on E-MMD (VG); but all the gaps are small. 

{\bf Qualitative Results.}
Some qualitative results of \OurModel are shown in~\cref{fig:sgg} (more in~\cref{app:sgg_qual}). As can be seen, the generated \groundedscenegraphs{} are reasonable. On VG, \OurModel learns the sparsity of the semantic edges. While on COCO-Stuff, \OurModel captures the fully-connected graph pattern.

\begin{table*}[t]
    \caption{\textbf{Ablations} regarding the different categorical input representations and the IoU loss on both the Visual Genome and COCO-Stuff validation sets. The model without the IoU loss is denoted as \OurModelnoIOU. 
    The evaluation metrics are the same as the ones used in~\cref{tab:sgg_vg}.
    }
    \centering
    \resizebox{\linewidth}{!}{
    \begin{tabular}{l|ccccc|cccc}
    \toprule 
    \multicolumn{10}{c}{Visual Genome (VG)} \\
    \midrule
    Method & N-MMD$\downarrow$ & F1-V$\uparrow$ & F1-A$\uparrow$ & F1-F$\uparrow$ & F1-BO$\uparrow$ & D-MMD$\downarrow$ & E-MMD$\downarrow$ & T-TV (val)$\downarrow$ & T-TV (train)$\downarrow$ \\ 
    \midrule
    \OurModel (bit) & 6.64e-3 & \textbf{0.184} & \textbf{0.308} & \textbf{0.292} &{0.747} & {5.26e-3} & 3.46e-2 & \textbf{0.702} & \textbf{0.685} \\
    \OurModel (scalar) & \textbf{4.91e-3} & 0.109 & 0.210 & 0.262 & 0.742 & \textbf{4.91e-3} & 7.61e-2 &  0.982 & 0.824 \\
    \OurModel (one-hot) & 2.94e-2 & 0.108 & 0.196 & 0.260 & \textbf{0.802} & 1.84e-2 & \textbf{2.96e-2} & 0.978 & 0.825 \\
    \midrule
    \OurModelnoIOU (bit) & {6.57e-3} & \textbf{0.173} & \textbf{0.285} & \textbf{0.283} & \textbf{0.736} & {7.85e-3} & \textbf{3.40e-2} & \textbf{0.709} & \textbf{0.692} \\
    \OurModelnoIOU (scalar) & 8.65e-3 & 0.168 & 0.267 & 0.276 & 0.712 & \textbf{7.69e-3} & 4.77e-2 & 0.729 & 0.713 \\
    \OurModelnoIOU (one-hot) & \textbf{3.05e-3} & 0.142 & 0.249 & 0.253 & 0.689 & 8.94e-3 & 5.77e-2 & 0.795 & 0.751 \\
    \bottomrule
    \toprule
    \multicolumn{10}{c}{COCO-Stuff} \\
    \midrule
    Method & N-MMD$\downarrow$ & F1-V$\uparrow$ & F1-A$\uparrow$ & F1-F$\uparrow$ & F1-BO$\uparrow$ & D-MMD$\downarrow$ & E-MMD$\downarrow$ & T-TV (val)$\downarrow$ & T-TV (train)$\downarrow$  \\ 
    \midrule
    \OurModel (bit) & \textbf{5.53e-4} & \textbf{0.421} & \textbf{0.485} & \textbf{0.637} & \textbf{0.830} & 0 & {7.25e-5} & \textbf{0.270} & \textbf{0.219} \\
    \OurModel (scalar) & 1.24e-2 & 0.172 & 0.222 & 0.368 & 0.822 & 0 & \textbf{1.41e-5} & 0.895 & 0.915 \\
    \OurModel (one-hot) & {1.86e-2} & 0.161 & 0.206 & 0.317 & 0.824 & 0 & 2.89e-3 & 0.807 & 0.793 \\
    \midrule
    \OurModelnoIOU (bit) & \textbf{5.63e-4} & \textbf{0.422} & \textbf{0.481} & \textbf{0.634} & \textbf{0.821} & 0 & \textbf{7.94e-5} & \textbf{0.272} & \textbf{0.225} \\
    \OurModelnoIOU (scalar) & 8.72e-4 & 0.380 & 0.432 & 0.605 & 0.788 & 0 & 4.65e-4 & 0.312 & 0.282 \\
    \OurModelnoIOU (one-hot) & 2.35e-3 & 0.365 & 0.372 & 0.553 & 0.762 & 0 & 1.82e-3 & 0.439 & 0.332 \\
    \bottomrule
    \end{tabular}
    }
    \label{tab:sgg_ablation}
\end{table*}

{\bf Ablations.} We conduct ablations with different categorical input representations (binary-bit, scalar, and one-hot), both with and without the IoU loss~(\cref{eq:loss_iou}). \cref{tab:sgg_ablation} shows the ablation results, where the model without the IoU loss is denoted as \OurModelnoIOU. Comparing among the three different categorical input encoding methods on both \OurModel and \OurModelnoIOU, the binary-bit representation works the best on both Visual Genome and COCO-Stuff datasets. Comparing \OurModel (bit) with \OurModelnoIOU (bit) on the F1 scores, we can see the effectiveness of our proposed IoU loss. The IoU loss seems not working well while using scalar and one-hot encodings. We believe that with these two types of representations the hyperparameters need to be further tuned. Current set of hyperparameters is tuned using the binary-bit representation. Overall, using the binary-bit representation with the IoU loss gives us the best performance on both datasets.

\subsubsection{\GroundedSceneGraph{} Completion}
Our \OurModel is versatile; besides doing the pure \groundedscenegraph{} generation, it can also achieve a variety of \groundedscenegraph{} completion tasks. 
We mainly follow~\citet{lugmayr2022repaint} for all the completions. 
All completion tasks are conducted on VG. 

\begin{table}[t]
    \caption{\textbf{\Groundedscenegraph{} completion results} on VG validation set.}
    \centering
    \resizebox{\linewidth}{!}{
    \begin{tabular}{l|cccc|c|cccc|c}
    \toprule
    & \multicolumn{5}{c|}{Single Node Label Completion} 
    & \multicolumn{5}{c}{Single Edge Label Completion} \\
    Method & HR@1 & HR@10 & HR@50 & HR@100 & mA & HR@1 & HR@10 & HR@25 & HR@50 & mA \\
    \midrule
    DiGress & 12.4 & 21.2 & 65.3 & 91.2 & 20.7 & 9.8 & 30.9 & 41.4 & 63.2 & 15.8 \\
    \OurModel & \textbf{13.9} & \textbf{25.7} & \textbf{73.2} & \textbf{94.5} & \textbf{23.6} & \textbf{10.1} & \textbf{35.7} & \textbf{46.2} & \textbf{65.3} & \textbf{19.4} \\
    \bottomrule
    \end{tabular}
    }
    \label{tab:node_complete}
\end{table}

\begin{figure}[t]
  \centering
   \includegraphics[width=\linewidth]{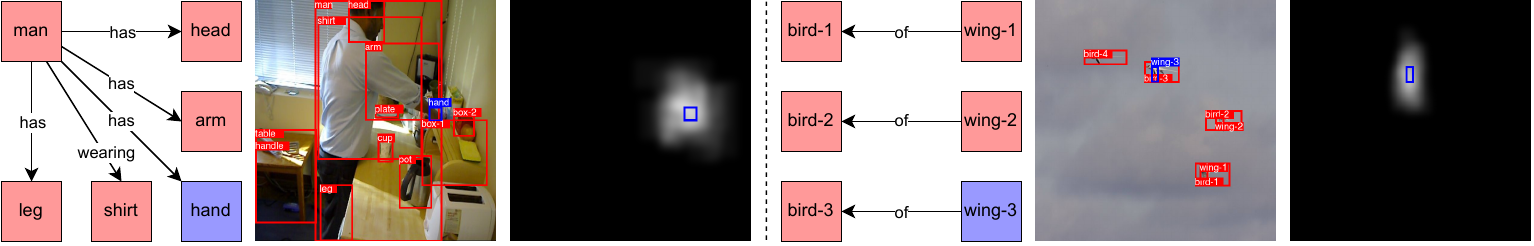}
   \caption{{\bf Single bounding box completion.}
   The left figure shows the input \groundedscenegraph, where only the edges and corresponding node labels are shown. The blue node's bounding box has been masked out.
   The middle figure shows the untouched (input) bounding boxes with labels in red, the one masked out 
   in blue, along with the corresponding ground-truth image. 
   The right figure shows our generated bounding box heatmap in white along with the target ground-truth bounding box (to be completed) in blue. The heatmap is obtained via generating the bounding box $100$ times; the whiter the area, the more overlap at the location.
   }
   \label{fig:complete}
\end{figure}

\textbf{Single Node Label Completion.} In this setting, 
we randomly masked out one node label per \groundedscenegraph{} in the validation set. 
For each \groundedscenegraph, we keep all the bounding box locations and edge labels, and let the model complete one node label given the remaining ones. The node whose label is masked out 
has degree (sum of in-degree and out-degree) at least $1$.  
This random masking is only done once, that is, 
it is fixed when evaluating the models. For each validation \groundedscenegraph, we conduct the completion $200$ times, and report the Hit Rate @ K (\textbf{HR@K}) and mean accuracy (\textbf{mA}) values. 
For each 
masked \groundedscenegraph, to calculate the HR@K, we first build a node label histogram from the $200$ completions over the $150$ object categories, and then keep the predictions from the K most frequently predicted node categories. We assign a score of $1$ if there is one prediction from the kept prediction set matches the ground-truth node label, and a score of $0$ otherwise. If there are multiple categories on the boundary when selecting the top K predicted categories, we randomly select some of them to make it exact K categories. Accuracy is defined as the ratio of the correct predictions (matched to the ground-truth) among the $200$ predictions. We then respectively take the average value of HR@K and accuracy over the whole validation set to get the final validation HR@K and mA scores. \cref{tab:node_complete} shows the HR@1/10/50/100 and mA results on the single node label completion task. As the results suggested, on all evaluation metrics, our \OurModel is consistently better than the DiGress~\citep{vignac2022digress} baseline. 

\textbf{Single Edge Label Completion.} Similar to the above single node label completion task, we conduct another single edge label completion task, where one edge label is masked out per validation \groundedscenegraph. The experiment setting and evaluation metrics are the same as the single node label completion setting. 
For each partially masked \groundedscenegraph, we predict the edge label $200$ times.
\cref{tab:node_complete} shows the HR@1/10/25/50 and mA results. 
Again, our \OurModel is consistently better than DiGress on all evaluation metrics. 

\textbf{Single Bounding Box Completion.} 
Some qualitative examples from \OurModel on the single bounding box completion task are shown in~\cref{fig:complete} (more in~\cref{app:box_complete}), where one bounding box location is masked out given a 
validation \groundedscenegraph, with all other information untouched. The node whose bounding box is masked out has degree at least $1$. 
Note that neither the image nor any image feature is given to the model 
for the completion task; the image is only for visualization. As seen, \OurModel can complete the bounding box in reasonable locations.

\subsection{Conditional Image Generator Discussion}
As described in~\cref{sec:model_images}, we have two conditional image generators: a layout-to-image model LayoutDiffusion~\citep{zheng2023layoutdiffusion} and a relation-aware model Relation-ControlNet. In this section, we are going to compare the image generation quality between these two generators. The COCO-Stuff dataset only has spatial relation types, which are less semantically meaningful because these relations can be simply decided from relative object bounding box locations. Thus we only use the VG dataset for the comparison.

\begin{table}[t]
    \caption{\textbf{\Groundedscenegraph{} classification evaluation results} in the PredCls and SGCls settings.}
    \centering
    \begin{tabular}{l|c|cc|c}
    \toprule
    & PredCls
    & \multicolumn{2}{c|}{SGCls} & \\
    Method & Mean Triplet Acc $\uparrow$ & Object Acc $\uparrow$ & Mean Triplet Acc $\uparrow$ & FID $\downarrow$ \\
    \midrule
    Ground-Truth Images & 30.65 & 70.31 & 15.77 & 0.0 \\
    \midrule
    LayoutDiffusion & \textbf{27.85} & \textbf{56.79} & \textbf{10.04} & 15.73 \\
    Object-ControlNet & 26.68 & 55.05 & 8.35 & \textbf{15.32} \\
    Relation-ControlNet & 27.08 & 47.94 & 9.07 & 15.99 \\
    \bottomrule
    \end{tabular}
    \label{tab:image_acc}
\end{table}

\subsubsection{\GroundedSceneGraph{} Classification Evaluation}
The \groundedscenegraph{} classification evaluation is to measure whether the generated images follow the condition control, that is, the object labels and locations, and the relation labels. For each generated image, we use trained \groundedscenegraph{} classification models to classify \groundedscenegraphs{} under two settings: (1) PredCls, where given the ground-truth object labels and bounding box locations, to classify the relation labels; and (2) SGCls, where given the object bounding box locations, to classify both object labels and relation labels. The PredCls setting is to evaluate whether the generated images contain the input relations at appropriate locations, while the SGCls setting is to check whether the objects and relations are generated properly as a whole at the given locations.

We use classification accuracies as the evaluation metrics. We calculate two types of accuracies: one for object labels, and one for triplet labels (the subject, relation, object labels in the <{\tt subject}, {\tt relation}, {\tt object}> triplets). Given the long-tailed nature of the relation labels in the VG dataset, while calculating the triplet label accuracy, we first take the average of the accuracies for each relation category, and then average across all relation categories, which weighs each relation category equally. The triplet accuracy is calculated under both the PredCls and SGCls settings while the object accuracy is calculated only under SGCls, since all the object information is given in PredCls.

The \groundedscenegraph{} classification models used are the MOTIFS-SUM-TDE models~\footnote{The model checkpoints are downloaded from \url{https://github.com/KaihuaTang/Scene-Graph-Benchmark.pytorch}.}~\citep{tang2020unbiased}, given their popularity, easy access, and reasonable performance. Besides LayoutDiffusion and Relation-ControlNet, we also consider another variant of ControlNet as a baseline, named Object-ControlNet. The differences between Object-ControlNet and Relation-ControlNet are that in Object-ControlNet, the control input contains only the object information and the text prompt includes all English words of object labels (allowing repetitions).
This can be considered as an ablation of Relation-ControlNet, which omits relation information. Note, both LayoutDiffusion and Object-ControlNet generate images conditioned on layout, while Relation-ControlNet generates images from \groundedscenegraphs.

Given an image generator, we use the $5,000$ \groundedscenegraphs{} from the VG's validation set, each to generate $5$ images, resulting in $25,000$ images. The accuracy results are presented in \cref{tab:image_acc}\footnote{As discussed in~\cref{app:sgg_layoutdiff}, the result calculation in this Table is actually inferior to LayoutDiffusion. However, it achieves the best accuracies compared to both Object-ControlNet and Relation-ControlNet.}, where the object label accuracy is denoted as \textbf{Object Acc} and the triplet label accuracy is denoted as \textbf{Mean Triplet Acc}. We also report the accuracies of the \groundedscenegraph{} classifiers on VG's ground-truth validation images, to show the upper bounds of the accuracy values. We also present the FID scores of the generated images to show the general image generation quality. 

As the results suggested, the images generated from all conditional image generators have comparable general image quality (similar FIDs). However, different image generators show different control abilities. Though both LayoutDiffusion and Object-ControlNet receive the same object control information: object labels and bounding box locations, the Object-ControlNet shows worse object renderings, which also results in worse relation renderings. We believe this result difference is due to the different model architectures and training strategies between these two models. The Relation-ControlNet includes relation labels as an additional control input compared to Object-ControlNet, thus it has better triplet accuracies even with worse object accuracy. The worse object renderings of Relation-ControlNet is reasonable. It is a common observation that it is more difficult to generate images properly with more condition inputs. 
Note that even the triplet accuracies of Relation-ControlNet are better than those of Object-ControlNet, the differences are marginal ($0.4$ under PredCls and $0.72$ under SGCls), and they are still worse than those of LayoutDiffusion. 
This is due to the fact that in VG, many relations can be inferred from bounding box positions. This is an artifact of the dataset and not a general problem at hand. 

\begin{table}[t]
    \caption{\textbf{Relation control evaluation results} under the PredCls and SGCls settings.}
    \centering
    \begin{tabular}{l|c|cc}
    \toprule
    & PredCls
    & \multicolumn{2}{c}{SGCls}\\
    Method & Mean Triplet Acc $\uparrow$ & Object Acc $\uparrow$ & Mean Triplet Acc $\uparrow$ \\
    \midrule
    LayoutDiffusion & 21.10 & \textbf{58.32} & 9.10 \\
    Object-ControlNet & 17.88 & 56.28 & 7.95 \\
    Relation-ControlNet & \textbf{24.23} & 55.21 & \textbf{10.95} \\
    \bottomrule
    \end{tabular}
    \label{tab:relation_acc}
\end{table}

\begin{figure*}[!ht]
  \centering
   \includegraphics[width=\linewidth]{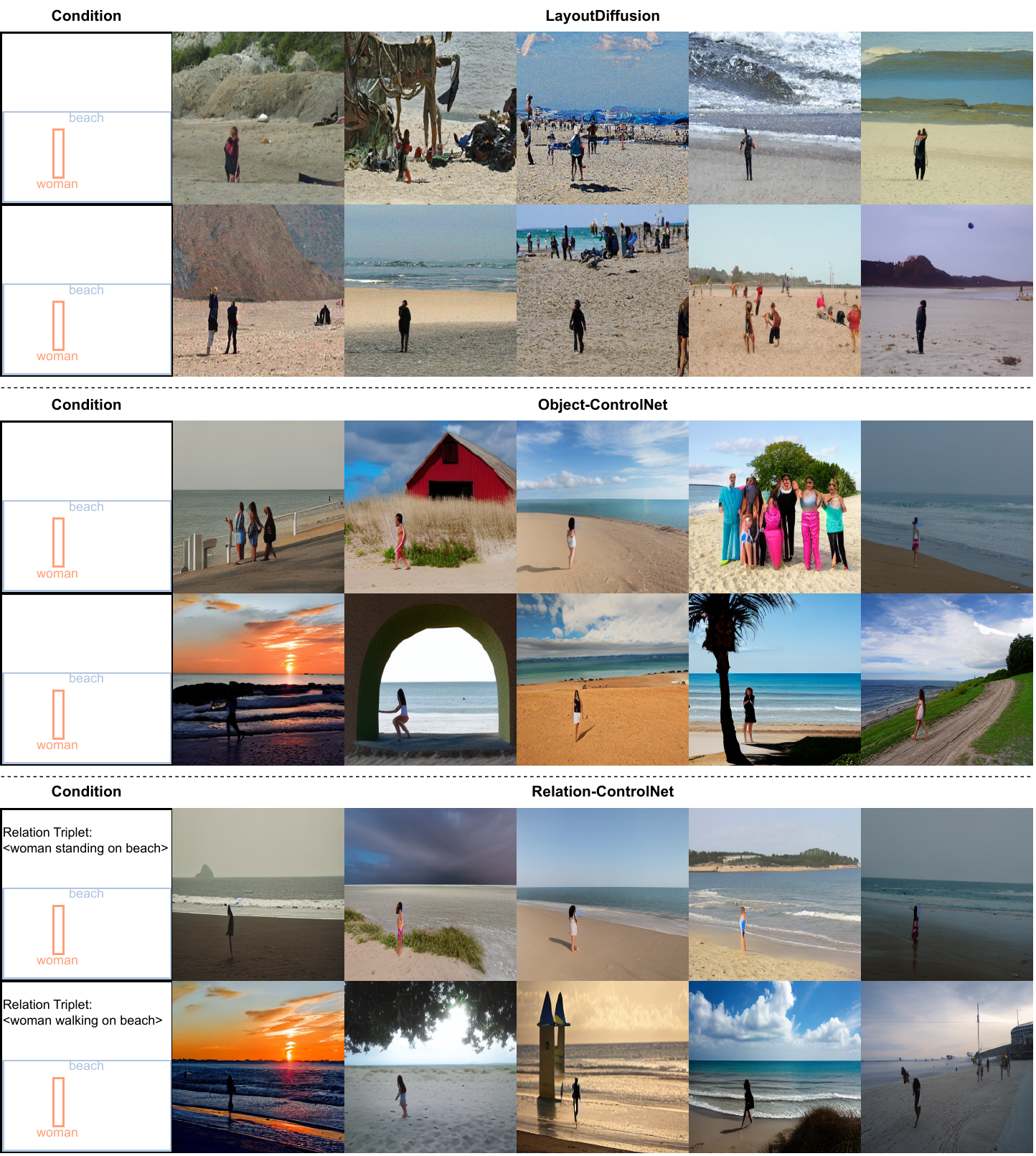}
   \caption{{\bf Relation control evaluation qualitative results.} The same layout condition containing an object {\tt woman} and an object {\tt beach} with their corresponding bounding box locations is fed to all the models. Additionally, the relation information is given to the Relation-ControlNet model. We draw $10$ image samples from both LayoutDiffusion and Object-ControlNet, while with Relation-ControlNet, we draw $5$ samples on the relations <{\tt woman standing on beach}> and <{\tt woman walking on beach}> respectively.}
   \label{fig:relation_exp}
\end{figure*}

\subsubsection{Relation Control Evaluation}
To show the benefits of Relation-ControlNet, we build an evaluation set specifically designed for relation control evaluation. Among the $50$ VG relation categories, we choose the subset containing {\tt carrying}, {\tt eating}, {\tt holding}, {\tt laying on}, {\tt looking at}, {\tt lying on}, {\tt playing}, {\tt riding}, {\tt sitting on}, {\tt standing on}, {\tt using}, and {\tt watching}. These are the relations which can not be easily determined by relative bounding box locations. In the validation set, we find the <{\tt subject}, {\tt relation}, {\tt object}> triplets containing those relation categories. For each such triplet, we find its closest triplet inside the validation set. The closest triplet is to have the same subject and object labels, but a different relation label as the original one. The relative bounding box position of the subject and the object of the closest triplet is also the same as the original one\footnote{The relative bounding box positions are defined as the same way used in the COCO-Stuff dataset.}. We discard the triplets for which we can not find the closest counterpart. If there are more than one such closest triplets, we choose the one which has the most similar subject and object bounding box aspect ratios as the original triplet. We then swap the relation labels between the triplet and its closest counterpart. We keep all the triplets, their closest triplets, and the triplets with swapped relations. This process gives us $328$ triplets. The swapping procedure guarantees that for the same pair of subject and object, there exist two different relation types, which is agnostic to the object-only conditioned models, but important to the relation conditioned model. For each triplet, we generate $5$ images from LayoutDiffusion, Object-ControlNet, and Relation-ControlNet respectively. The corresponding object and triplet accuracies are reported in~\cref{tab:relation_acc}.

This setting shows the benefits of Relation-ControlNet. As indicated in the results, though Relation-ControlNet is worse than LayoutDiffusion and Object-ControlNet in terms of object accuracy, its triplet accuracies are significantly better than the other two. This is because LayoutDiffusion and Object-ControlNet can not take the relation label as input but Relation-ControlNet can. As the qualitative examples shown in~\cref{fig:relation_exp}, when the relation information is inputted to the model, Relation-ControlNet can constantly render the appropriate postures for the {\tt woman}, either standing or walking depending on the condition. However, for both LayoutDiffusion and Object-ControlNet, since these models can not read in the relation information, the models have to render the postures of the {\tt woman} via their own decisions, which may be standing, walking, or even bending the knees. Though our Relation-ControlNet is not perfect, it implies the usefulness of generating \groundedscenegraphs. Generating \groundedscenegraphs, which contain object labels and locations along with relation labels, can provide more controllability for image generation. 

Relation-ControlNet is designed for a harder task, where the relation information is also part of the condition.
However, given the inferior performance of Relation-ControlNet in the real validation setting~(\cref{tab:image_acc}) and limitations of current datasets (many of the relations in VG can be easily determined by relative bounding box positions), we 
use LayoutDiffusion as the conditional image generator for experiments that follow.

\begin{table}[t]
  \caption{\textbf{\Groundedscenegraph{} detection (SGDet) results} of SGTR.}
  \centering
  \begin{tabular}{l|c|c|c|c|c|c|c}
    \toprule
    \textbf{Data Setting} & \textbf{mR@50} & \textbf{mR@100} & \textbf{R@50} & \textbf{R@100} & \textbf{Head} & \textbf{Body} & \textbf{Tail} \\
    \midrule
    Original & 11.9 & 16.0 & \textbf{23.7} & \textbf{27.0} & \textbf{26.6} & 19.8 & 9.0 \\
    Additional & \textbf{12.7} & \textbf{16.7} & 23.6 & 26.8 & 26.4 & \textbf{20.7} & \textbf{9.7} \\
    \bottomrule
  \end{tabular}
  \label{tab:sgtr_body}
\end{table}

\subsection{\GroundedSceneGraph{} Detection Evaluation}
We take the \OurModel trained on VG, 
and pair it with the trained LayoutDiffusion model~\citep{zheng2023layoutdiffusion} to form $5,000$ \groundedscenegraph{} - image pairs, treated as additional training data for the downstream \groundedscenegraph{} detection (SGDet) task: given an image, detecting the object labels and locations, as well as the relation labels. 
Those additional \groundedscenegraphs{} only contain body relations as defined in~\citet{li2021bipartite}. Detailed process is in~\cref{app:sgtr}.

We use the SGTR model~\citep{li2022sgtr}, as an example, to show the value of our generated \groundedscenegraph{} - image pairs. 
We train SGTR for the SGDet task under two training data settings: (1) Original, where the training data is the original Visual Genome training data in~\citet{li2022sgtr}; (2) Additional, where besides the original training data, we add in our generated $5,000$ \groundedscenegraph{} - image pairs. Note that these two settings only differ in the training data; both validation and testing data is still the original 
data for the SGDet task. For both data settings, we train SGTR 
$4$ times and the averaged test results are reported in~\cref{tab:sgtr_body}, where the model for testing is selected via best mR@100 on the validation set. 
We report 
results on mean Recalls (mR@50 and mR@100), Recalls (R@50 and R@100), and mR@100 on the head, body, and tail relation partitions (respectively indicated as Head, Body, and Tail in the Table). 

Comparing our Additional results with the Original ones, we can see that our generated \groundedscenegraph{} - image pairs do have value, which brings improved results on Body and Tail, and comparable results on Head, which results in increased results on mean Recalls and comparable results on Recalls. As suggested in~\citet{tang2020unbiased}, mean Recall is a better evaluation metric than Recall for the \groundedscenegraph{} detection task, because it is less biased to the dominant relation classes. Note that although the generated \groundedscenegraphs{} in the additional training data only contain body relations, the Tail results also get improved. This is reasonable, because \groundedscenegraph{} is a structure, increasing the confidence of some part of the structural prediction will increase the confidence of other part as well, especially for the tail relations, where the prediction confidence is usually low.

\section{Conclusion} 
In this work, we propose a novel 
framework for joint \groundedscenegraph{} - image generation. As part of this, 
we propose \OurModel, a diffusion-based model for generating \groundedscenegraphs{} that adeptly handles mixed discrete and continuous attributes. \OurModel demonstrates superior performance on both unconditional \groundedscenegraph{} generation and conditional \groundedscenegraph{} completion tasks. By pairing \OurModel with a conditional image generation model, the joint \groundedscenegraph{} - image pair distribution can be obtained. 
We illustrate the benefits of \OurModel both on its own and as part of joint \groundedscenegraph{} - image generation. 
In the future, we are interested in modeling the joint distribution with a single model.

\subsubsection*{Acknowledgments}
This work was funded, in part, by the Vector Institute for AI, Canada CIFAR AI Chairs, NSERC CRC, and NSERC DGs. Resources used in preparing this research were provided, in part, by the Province of Ontario, the Government of Canada through CIFAR, the Digital Research Alliance of Canada \url{alliance.can.ca}, \href{https://vectorinstitute.ai/\#partners}{companies} sponsoring the Vector Institute, and Advanced Research Computing at the University of British Columbia. Additional hardware support was provided by John R. Evans Leaders Fund CFI grant and Compute Canada under the Resource Allocation Competition award.
Qi Yan is supported by the UBC Four Year Doctoral Fellowship.

\bibliography{main}
\bibliographystyle{tmlr}

\newpage
\appendix
\section{Implementation Details}
\subsection{Discrete Input Encodings} \label{app:input_code}
To effectively handle the discrete attributes, we implement the following encoding methods.

{\bf Scalar.} 
Similar to prior works~\citep{ho2020denoising, song2021scorebased, karras2022elucidating, jo2022score}, we use zero-based indexing and map the scalar value to the range $[-1, 1]$. 
Specifically, during training, denoting an integer label of an attribute as $n$ and the number of categories of the attribute as $m$, where $n \in \{0, 1, \dots, m-1\}$, the scalar representation becomes $\frac{2n}{m-1}-1$.
During sampling, we first split the interval $[-1, 1]$ into equal-sized bins in accordance with the number of node or edge categories. We then decode the continuous-valued network output into a discrete label based on the bin into which the output value falls.

{\bf Binary-Bit.} 
Following~\citet{chen2022analog}, we first convert the zero-based indexed integer node/edge attribute into binary bits, and then remap the 0/1 bit values to -1/1 for improved training stability. During sampling, we binarize each value in the network output based on its sign. That is, a positive value is interpreted as $1$, and a negative value as $0$. We then convert the binary representation back into an integer node or edge label.

{\bf One-Hot.} 
We remap the 0/1 values in the one-hot encoding of the original integer node/edge label to -1/1. During sampling, we take the $\operatorname{argmax}$ value of the network output to obtain the categorical label.

\subsection{\OurModel Network Architecture} \label{app:diffusesg_net}
Our proposed {\em graph transformer} has  
two essential components: (1) shifted-window attention mechanism, and (2) downsampling/upsampling layers. In the case of graphs comprising \(n\) nodes, their adjacency matrices, incorporating both node and edge attributes, are conceptualized as high-order tensors with \(n \times n\) entries. To handle graphs of varying sizes, we standardize the size of these adjacency matrices through padding. Consequently, for different datasets, we accordingly adjust the design parameters to ensure the network is proportionate and suitable for the specific requirements of each dataset.

\textbf{Shifted-Window Attention.}
We adopt the shifted window attention technique from~\citet{liu2021swin}, which partitions the original grid-like feature map into smaller subregions. Within these subregions, local message passing is executed using self-attention mechanisms. Additionally, the windows are interleavedly shifted, facilitating cross-window message passing, thereby enhancing the overall efficiency and effectiveness of the feature extraction process.

\begin{figure*}[t]
  \centering
   \includegraphics[width=0.75\linewidth]{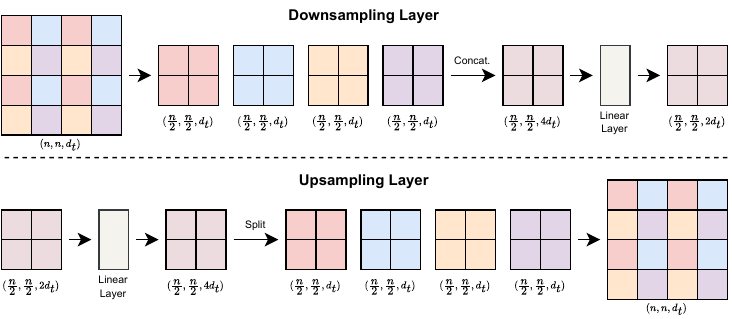}
   \caption{{\bf Visualization} of the downsampling and upsampling layers.}
   \label{fig:downup}
\end{figure*}

\textbf{Downsampling/Upsampling Layer.}
We incorporate channel mixing-based downsampling/upsampling operators to effectively diminish or augment the size of the feature map, thereby constructing hierarchical representations. 
During downsampling, the feature map is divided into four segments based on the parity of the row and column indices, followed by a concatenation process along the channel, which serves to reduce the dimensions of height and width. 
The upsampling process performs the inverse operations. It initially splits the tensors along the channel and then reshapes them, effectively reversing the process conducted in downsampling.
The downsampling and upsampling layers are visualized in~\cref{fig:downup}.
We also implement one MLP layer right after each downsampling/upsampling layer.
In line with the widely recognized U-Net architecture~\citep{song2021scorebased,karras2022elucidating}, our approach also integrates skip-connections for tensors of identical sizes to enhance the network capacity. 

The crucial design parameters of our model are detailed in~\cref{tab:net_details}. It is important to note that within the Down/Up block layers, the initial blocks do not utilize downsampling/upsampling operations. For instance, in the context of the Visual Genome dataset, we effectively implement $3$ downsampling layers, which leads to the successive alteration of the feature map dimensions as \(64 \rightarrow 32 \rightarrow 16 \rightarrow 8\). In this setup, we opt for a window size of $8$, ensuring that the receptive field is sufficiently large to facilitate effective message passing between each pair of nodes. 
While on COCO-Stuff, we employ $2$ downsampling layers, resulting the feature map dimensions as \(40 \rightarrow 20 \rightarrow 10\), and thus the window size is set to $10$.

\begin{table}[t]
\caption{
\textbf{Architecture details} of our graph transformer.
}
\centering
\begin{tabular}{l|cc}
\toprule
Hyperparameter & VG & COCO-Stuff\\ 
\midrule
Full tensor size  ($n \times n $) & 64 $\times$ 64 & 40 $\times$ 40 \\
Down block attention layers & [1, 1, 3, 1] & [1, 2, 6]\\
Up block attention layers & [1, 1, 3, 1] & [1, 2, 6] \\
Number of attention heads & [3, 6, 12, 24] & [3, 6, 12] \\
Window size & 8 & 10 \\
Token dimension & 96 & 96 \\
Feedforward layer dimension & 384 & 384 \\
\bottomrule
\end{tabular}
\label{tab:net_details}
\end{table}

{\bf MLP Prediction Head.} 
The node/edge attribute MLP prediction head 
is implemented as two linear layers with a $\operatorname{GELU}$~\citep{hendrycks2016gaussian} operation injected in between.

\subsection{\OurModel Diffusion Modeling Details} \label{app:diffusesg_detail}
To ensure the stable training of our diffusion model, we adopt a framework based on the stochastic differential equation (SDE), as proposed in~\citet{song2021scorebased}. Additionally, we incorporate a variety of training techniques that have been proven effective in image generation contexts: network preconditioning~\citep{karras2022elucidating}, self-conditioning~\citep{chen2022analog} and exponential moving average (EMA). 
For network training, we employ the hyperparameters specified for the ImageNet-64 dataset in~\citet{karras2022elucidating} for preconditioning purposes; 
a detailed explanation of these parameters can be found therein. 
We use Adam optimizer and learning rate being $0.0002$. 
The EMA coefficients used for evaluation 
are $0.9999$ and $0.999$ on the Visual Genome and COCO-Stuff datasets respectively.

\begin{algorithm}[t]
\caption{\OurModel Sampler.}
\label{alg:sampler}
\begin{algorithmic}[1]
\REQUIRE{$D_\theta, T, \{t_i\}_{i=0}^T, \{\gamma_i\}_{i=0}^{T-1}.$}
\STATE {\bfseries sample} {$ \bwtS^{(0)} \sim \mathcal{N}(\boldsymbol{0}, t_0^2\bI), \bwhS^{(0)}_{\text{sc}}=\boldsymbol{0}$}.
\FOR{$i=0$ to $T-1$}
\STATE {\bfseries sample} {$\beps \sim \mathcal{N}(\boldsymbol{0}, S_{\text{noise}}^2\bI)$} 
\STATE {$ \hat{t}_{i} \leftarrow (1 + \gamma_i) t_i$}
\STATE {$\bwtS^{(\hat{i})} \leftarrow \bwtS^{(i)} + \sqrt{\hat{t}_i^2 - t_i^2} \beps$}
\STATE {$\bwhS^{(\hat{i})}_{\text{sc}} \leftarrow  D_\theta(\bwtS^{(\hat{i})}, \bwhS^{(i)}_{\text{sc}},  \hat{t}_i)$}
\STATE {$\boldsymbol{d}_i \leftarrow (\bwtS^{(\hat{i})} - \bwhS^{({\hat{i}})}_{\text{sc}}) / \hat{t}_i$}
\STATE {$\bwtS^{(i+1)} \leftarrow \bwtS^{(\hat{i})} + (t_{i+1} - \hat{t}_i) \boldsymbol{d}_i$}
\STATE {$ \bwhS^{(i+1)}_{\text{sc}} \leftarrow D_\theta(\bwtS^{(i+1)}, \bwhS^{({\hat{i}})}_{\text{sc}},  {t}_{i+1})$}
\STATE { $\boldsymbol{d}_i' \leftarrow (\bwtS^{(i+1)} - \bwhS^{(i+1)}_{\text{sc}} ) / {t}_{i+1}$}
\STATE {$ \bwtS^{(i+1)} \leftarrow \bwtS^{(i)} + \frac{1}{2}(t_{i+1} - \hat{t}_i) (\boldsymbol{d}_i + \boldsymbol{d}_i')$}
\ENDFOR
\STATE {\bfseries return} $\bwtS^{(T)}$
\end{algorithmic}
\end{algorithm}

\begin{table}[t]
    \caption{\textbf{Sampling parameters} in the denoising process.}
    \centering
    \begin{tabular}{c c}
    \toprule
    \multicolumn{2}{c}{$\sigma_{\text{min}}=0.002, \sigma_{\text{max}}=80, \rho=7$} \\
    \multicolumn{2}{c}{$S_\text{tmin}=0.05, S_{\text{tmax}}=50, S_{\text{noise}}=1.003, S_{\text{churn}}=40, T=256$} \\
    \multicolumn{2}{c}{$t_i = ({\sigma_{\text{max}}}^\frac{1}{\rho} + \frac{i}{T-1}({\sigma_{\text{min}}}^\frac{1}{\rho} - {\sigma_{\text{max}}}^\frac{1}{\rho}))^\rho$} \\
    \multicolumn{2}{c}{$\gamma_i = \boldsymbol{1}_{S_\text{tmin} \leq t_i \leq S_\text{tmax}} \cdot \min(\frac{S_{\text{churn}}}{T}, \sqrt{2}-1)$} \\
    \bottomrule
    \end{tabular}
    \label{tab:sampler_hp}
\end{table}

The pseudocode of our sampling algorithm is presented in~\cref{alg:sampler}, 
which follows the stochastic sampler in~\citet{karras2022elucidating} but with the additional self-conditioning~\citep{chen2022analog} technique. 
In the algorithm, $D_\theta$ is the denoising network, $\bwtS^{(t)}$ is the generated \groundedscenegraph{} at step $t$, and $\bI$ is the identity matrix. 
The associated parameters are detailed in~\cref{tab:sampler_hp}. 
We opt for \(T=256\) sampling steps to expedite the sampling process, as opposed to the original \(1,000\) steps used in the DDPM framework~\citep{ho2020denoising}.
In~\cref{alg:sampler}, with a slight abuse of notations for simplicity, we consider the generated \groundedscenegraph{} ($\bwtS$, $\bwhS_{\text{sc}}$), which comprises tuples of node and edge attributes, as a singular tensor, allowing for straightforward addition or subtraction operations. Practically, this is implemented through separate operations on the node and edge tensors.

\subsection{Relation-ControlNet and Object-ControlNet Training Details} \label{app:controlnet}
Following~\citet{zhang2023adding}, we use Stable Diffusion as an instantiation of the ControlNet architecture. We use Stable Diffusion V1.5. To train both Relation-ControlNet and Object-ControlNet, we use Adam optimizer with $\beta_1$ being $0.9$, $\beta_2$ being $0.999$, and weight decay being $0.01$; a constant learning rate $0.00001$ is used to train the models. Both models are trained for $200$ epochs with a batch size of $120$. Following~\citet{zhang2023adding}, during training, the text prompts are randomly replaced with empty strings at a chance of $50\%$. Images are generated in the resolution of $256 \times 256$.

\section{D3PM Baseline} \label{app:d3pm}
Our D3PM~\citep{austin2021structured} baseline is based on the image generation model on the CIFAR-10 dataset~\citep{krizhevsky2009learning}. 
Given a scene graph $\boldsymbol{S} = (\boldsymbol{V}, \boldsymbol{E})$\footnote{We slightly abuse the notations here, specifically for the D3PM model, compared to the ones in the main text.} with $n$ nodes, where $\boldsymbol{V} \in \mathbb{N}^{n}$ is the node vector containing the integer node labels, and $\boldsymbol{E} \in \mathbb{N}^{n \times n}$ is the adjacency matrix containing the integer edge labels, the input to our D3PM baseline is represented as $\boldsymbol{Q} \in \mathbb{N}^{n \times n \times 3}$, where $\boldsymbol{Q}_{i, j} = [\boldsymbol{V}_i, \boldsymbol{V}_j, \boldsymbol{E}_{i, j}]$, $\forall i, j \in \{1, 2, \dots, n\}$. 

We use two separate discretized Gaussian transition matrices, one for the node category and one for the edge category, to add noise on $\boldsymbol{Q}$, resulting in the noised $\boldsymbol{\tilde{Q}}$. 
We then use a U-Net with two separate prediction heads, each implemented as two convolution layers with a $\operatorname{sigmoid}$ operation before each of the convolution layers, to respectively produce the logits of the denoised $\boldsymbol{\hat{V}}$ and $\boldsymbol{\hat{E}}$, which then form the logits of the denoised $\boldsymbol{\hat{Q}}$. 
We use the $L_{\lambda=0.001}$ (Eq.~(5) in~\citet{austin2021structured}), calculated on the logits of $\boldsymbol{\hat{Q}}$, to train our D3PM baseline. 
We use Adam optimizer and learning rate being $0.00005$ for training. We use $T = 1,000$ noising and denoising steps and take the model with EMA coefficient $0.9999$ for evaluation. 

The $\beta_t$ (in Eq.~(8) in the Appendix of~\citet{austin2021structured}) of the discretized Gaussian transition matrix is increased linearly, for $t \in \{1, 2, \dots, T\}$.
On the Visual Genome dataset, we set $n$ to be $64$, and $\beta_t$ is increased linearly from $0.0001$ to $0.02$ for both node and edge categories. On the COCO-Stuff dataset, $n$ is set to be $36$. The $\beta_t$ is increased linearly from $0.0001$ to $0.02$ for the node category and from $0.04$ to $0.1$ for the edge category.

\section{Evaluation Details}
\subsection{\GroundedSceneGraph{} Classification Evaluation on LayoutDiffusion} \label{app:sgg_layoutdiff}
The LayoutDiffusion model checkpoint that we used is trained on a version of the VG dataset annotation which has $178$ object categories. This is slightly different from the VG dataset annotation that the \groundedscenegraph{} classification models are trained on, which contains $150$ object categories. However, between these two versions of VG annotations, there are $131$ object categories in common. Thus when generating images from LayoutDiffusion for the \groundedscenegraph{} classification evaluation, we only keep the objects whose labels are in the common category set. Specifically, given a ground-truth VG validation \groundedscenegraph, we keep the objects in the common category set and discard others, and then generate the corresponding images. But when calculating the \groundedscenegraph{} classification accuracy scores, we still use the ground-truth \groundedscenegraph{} without any object filtering. Though this setting is inferior to LayoutDiffusion, it still achieves better accuracies than Object-ControlNet and Relation-ControlNet, as shown in~\cref{tab:image_acc}.

When building the evaluation set for the relation control evaluation, we make sure that all the subject and object labels are in the common category set.

\subsection{Generating Additional Training Data for the \GroundedSceneGraph{} Detection Task} \label{app:sgtr}
We take our \OurModel model trained on the Visual Genome dataset, let it generate a set of \groundedscenegraphs{} which only contain relations falling into the body partition (as defined in~\citet{li2021bipartite}), and then use the pretrained (on VG, with resolution $256 \times 256$) LayoutDiffusion model~\citep{zheng2023layoutdiffusion} to form the \groundedscenegraph{} - image pairs. Since there exists some node label set discrepancy between the respective VG annotations used to train the LayoutDiffusion model ($178$ node categories) and our \OurModel model ($150$ node categories). When forming the \groundedscenegraph{} - image pairs, we discard the nodes whose labels are not in the common category set ($131$ node categories) and their related edges. We randomly choose $5,000$ such generated pairs, where node numbers are restricted to be less than $10$, as additional training data to train the SGTR model~\citep{li2022sgtr} on the \groundedscenegraph{} detection task: 
given an image, detecting a \groundedscenegraph{} (node labels and bounding box locations, and edge labels) from it. We guarantee that for those randomly chosen \groundedscenegraphs, each of them has at least one edge.

The motivations of why we generating the \groundedscenegraphs{} only containing body relations are as follows. First, for the \groundedscenegraph{} detection task, there are already many training instances for the head classes, so generating additional head relations may not be beneficial at all. Second, for our \groundedscenegraph{} generation task, since the training data for the tail relations is limited, our \groundedscenegraph{} generation model may not be able to model the tail class distribution well.

\section{More Qualitative Results}
\subsection{\GroundedSceneGraph{} - Image Pair Generation} \label{app:sgg_qual}
\begin{figure*}[t]
  \centering
   \includegraphics[width=0.7\linewidth]{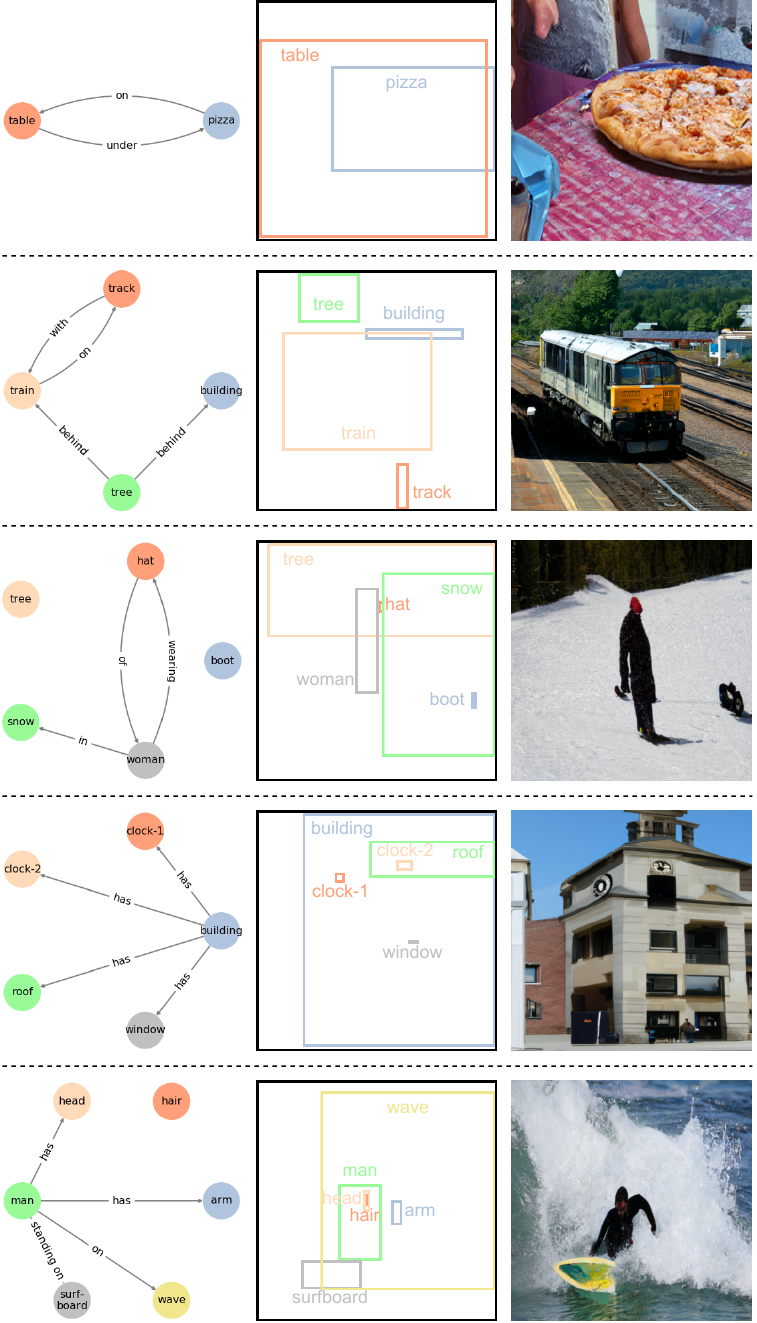}
   \caption{{\bf \Groundedscenegraph{} - image pair generation} qualitative results on the Visual Genome dataset.}
   \label{fig:sgg-vg1}
\end{figure*}

\begin{figure*}[t]
  \centering
   \includegraphics[width=0.7\linewidth]{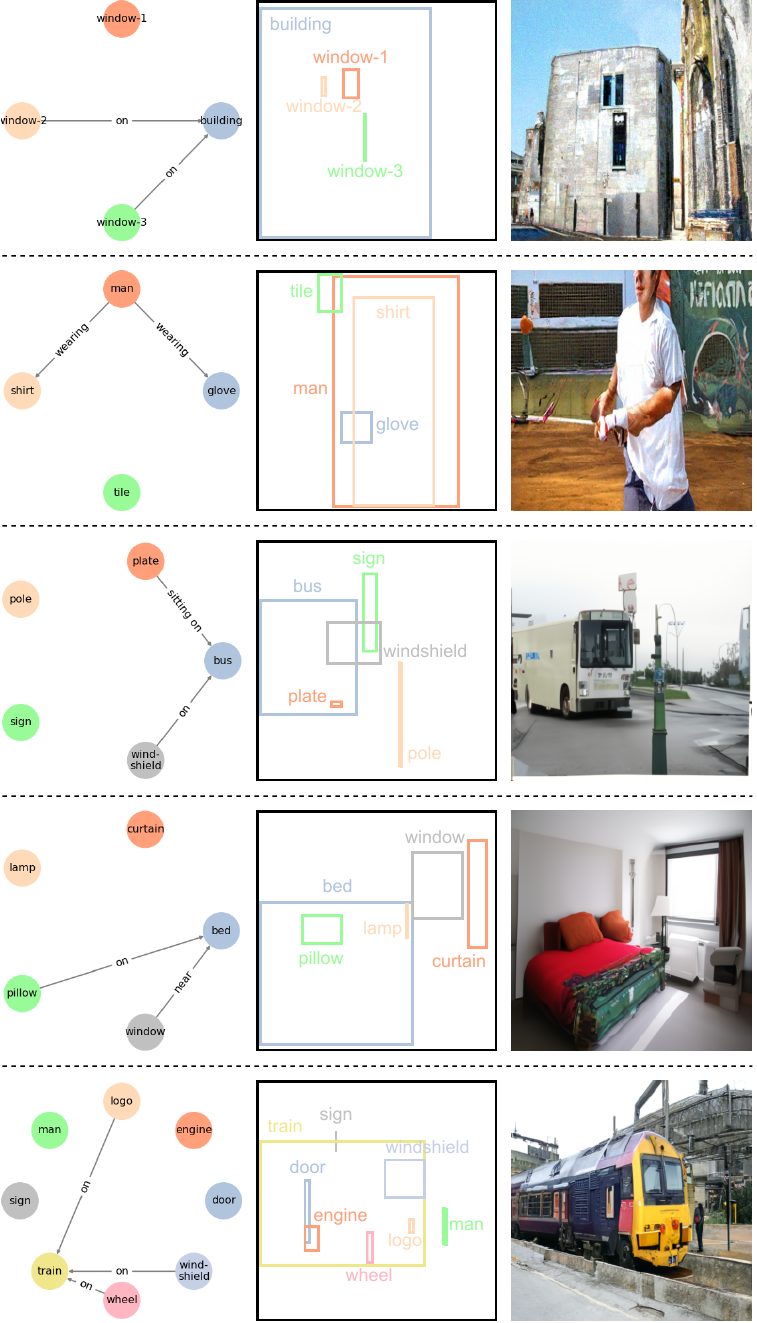}
   \caption{{\bf \Groundedscenegraph{} - image pair generation} qualitative results on the Visual Genome dataset.}
   \label{fig:sgg-vg2}
\end{figure*}

\begin{figure*}[t]
  \centering
   \includegraphics[width=0.8\linewidth]{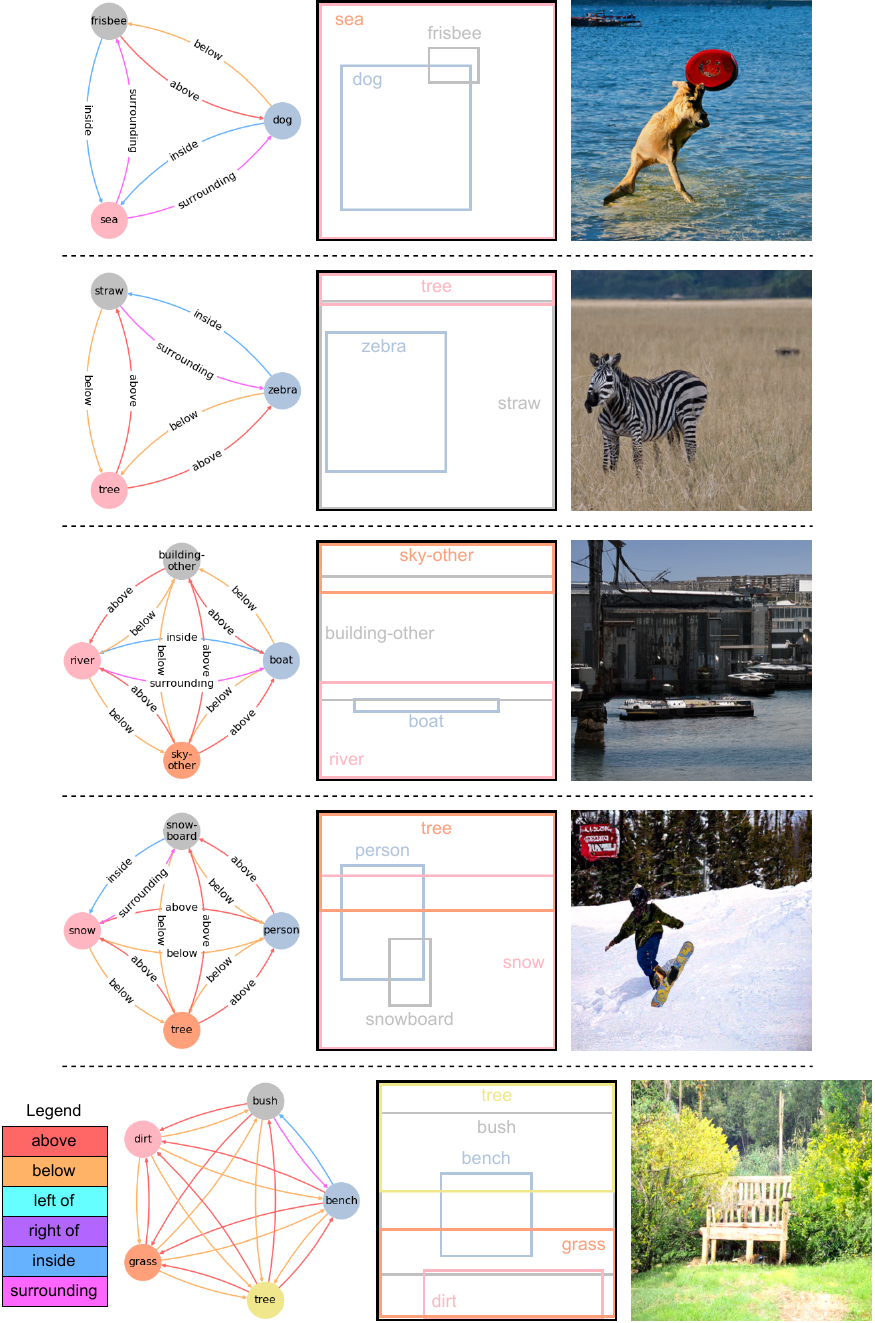}
   \caption{{\bf \Groundedscenegraph{} - image pair generation} qualitative results on the COCO-Stuff dataset.}
   \label{fig:sgg-coco1}
\end{figure*}

\begin{figure*}[t]
  \centering
   \includegraphics[width=0.8\linewidth]{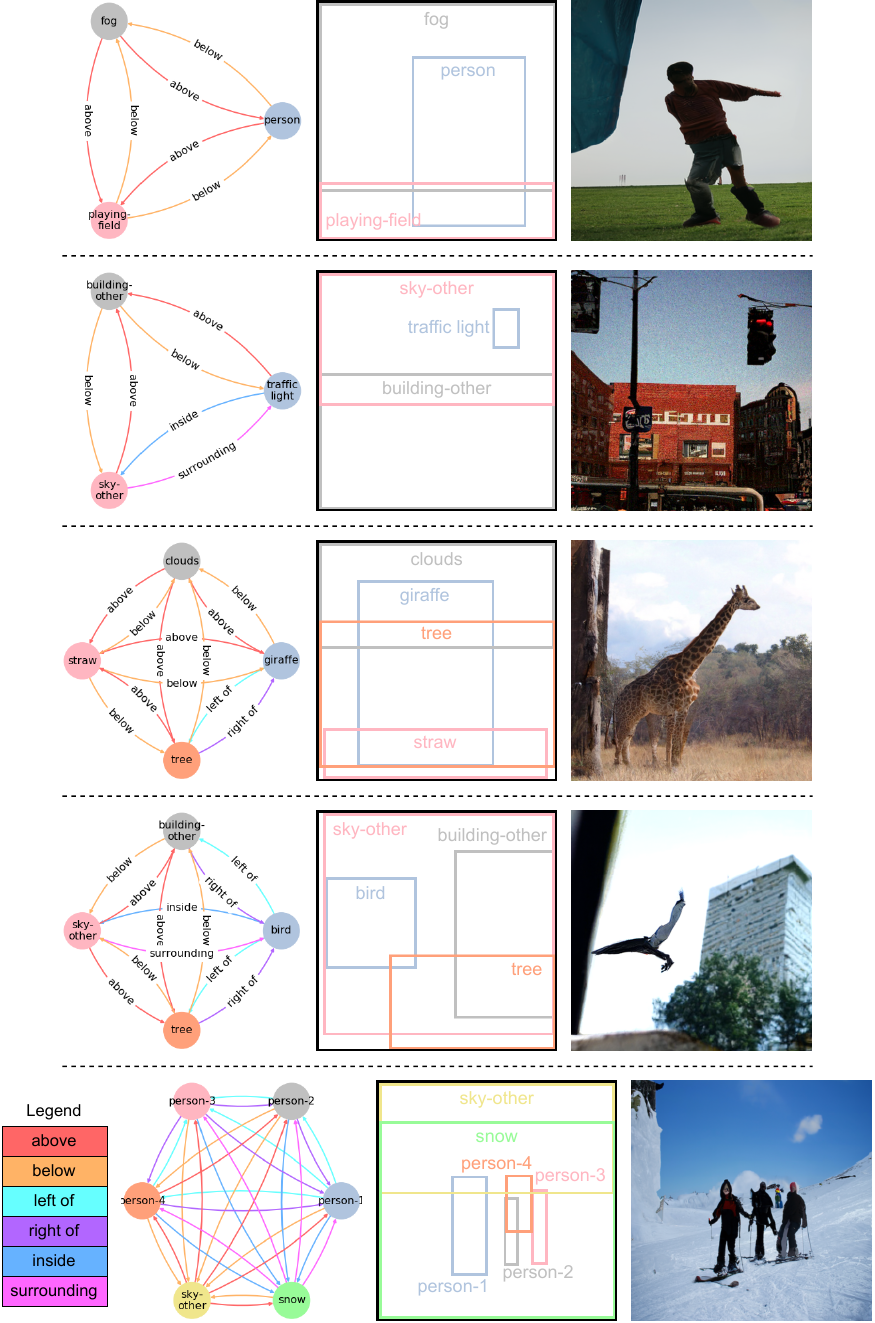}
   \caption{{\bf \Groundedscenegraph{} - image pair generation} qualitative results on the COCO-Stuff dataset.}
   \label{fig:sgg-coco2}
\end{figure*}

More qualitative results of \groundedscenegraph{} - image pair generation are shown in 
\cref{fig:sgg-vg1,fig:sgg-vg2} (Visual Genome) and \cref{fig:sgg-coco1,fig:sgg-coco2} (COCO-Stuff). 
\Groundedscenegraphs{} including the bounding box locations are generated by \OurModel and the corresponding images (in resolution $256 \times 256$) are produced by the pretrained LayoutDiffusion model~\citep{zheng2023layoutdiffusion}. 
Note that on the Visual Genome dataset, since there exists some node label set discrepancy between the annotations used to train our \OurModel model and the LayoutDiffusion model, we only visualize the \groundedscenegraphs{} whose node labels are all in the common node label set ($131$ node categories).

\subsection{Single Bounding Box Completion} \label{app:box_complete}
\begin{figure*}[t]
  \centering
   \includegraphics[width=0.7\linewidth]{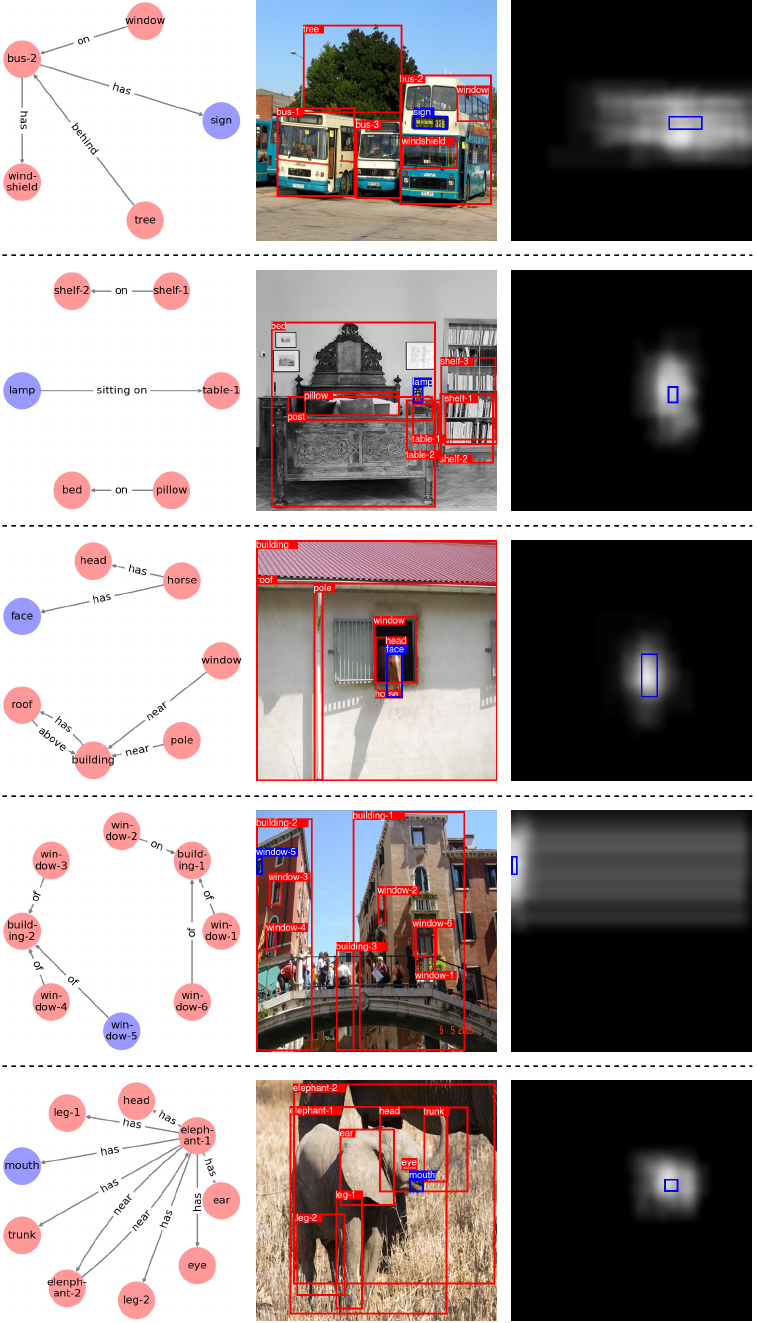}
   \caption{{\bf Single bounding box completion} qualitative results on the Visual Genome validation set.}
   \label{fig:bbox}
\end{figure*}

More qualitative results of our \OurModel on the Visual Genome validation set  
are shown in~\cref{fig:bbox}. 
The left figure shows the input \groundedscenegraph, where only the edges and corresponding node labels are shown. The blue node's bounding box has been masked out.
The middle figure shows the untouched (input) bounding boxes with labels in red, the one masked out 
in blue, along with the corresponding ground-truth image. 
The right figure shows our generated bounding box heatmap in white along with the target ground-truth bounding box (to be completed) in blue. The heatmap is obtained via generating the bounding box $100$ times; the whiter the area, the more overlap at the location. 
Note that neither the image nor any image feature is given to the model
for the completion task; the image is only for visualization.

\end{document}